\documentclass[twoside,11pt]{article}

\usepackage[font=small,labelfont=bf]{caption}
\usepackage{needspace}

\usepackage{dnd}
\usepackage{times}
\usepackage{natbib}
\usepackage{chngcntr}
\usepackage{amsmath}
\usepackage{stmaryrd}
\usepackage{booktabs}
\usepackage{hyperref}
\usepackage{float}
\usepackage{tabularx}
\usepackage{multirow}
\usepackage{stfloats}
\usepackage{inconsolata}
\usepackage{wrapfig}
\usepackage{listings}
\usepackage{caption}
\usepackage{gb4e}
\noautomath
\usepackage{ragged2e}
\usepackage{varwidth}
\usepackage{soul}
\usepackage{bbm}

\usepackage{siunitx}
\usepackage{threeparttable}
\usepackage[T1]{fontenc}
\usepackage[utf8]{inputenc}
\usepackage{textcomp}

\usepackage[capitalize,nameinlink,noabbrev]{cleveref}
\creflabelformat{equation}{#2\textup{#1}#3}
\crefname{section}{\S}{\S\S}  % Like: § 4
\crefname{subsection}{\S}{\S\S}  % Like: § 4.1
\crefname{subsubsection}{\S}{\S\S} % Like: § 4.1.1

\crefname{codeboxinput}{Listing}{Listings}
\usepackage{fontawesome5}
\usepackage{algorithm}
\usepackage{algpseudocodex}
\usepackage{marginnote}

\usepackage[dvipsnames]{xcolor}
\newcommand{\edit}[2][]{%
  \textcolor{blue}{#2}%
  \ifx&#1&%
  \else
    \marginnote{\raggedright\tiny\textcolor{orange}{#1}}%
  \fi
}
\usepackage[most]{tcolorbox}
\tcbset {
  base/.style={
    arc=0mm,
    bottomtitle=0.5mm,
    boxrule=0mm,
    colbacktitle=black!10!white,
    colframe=black!30!white,
    coltitle=black,
    fonttitle=\bfseries\fontsize{10}{12}\selectfont,
    left=2.5mm,
    leftrule=1mm,
    right=3.5mm,
    title={#1},
    toptitle=0.75mm,
    breakable,
    listing options={style=tcblatex, breakindent=0pt, breaklines=true, basicstyle=\ttfamily\fontsize{8}{9}\selectfont}
  }
}

\newtcblisting[auto counter]{codebox}[2][]{
  base={Listing \thetcbcounter: #2},
  listing only,
  #1
}

\newtcbinputlisting[use counter=codeboxinput]{\codeboxinput}[3][]{%
  listing file={#3},
  base={Listing \thetcbcounter: #2},
  listing only,
  #1
}

\newtcolorbox{subbox}[2][]{
  colframe=black!30!white,
  base={#2},
  #1
}
\newcommand{\ben}{\colorbox{green!20}{\textsc{beneficial}}}
\newcommand{\detr}{\colorbox{red!20}{\textsc{detrimental}}}
\newcommand{\neu}{\colorbox{yellow!20}{\textsc{neutral}}}
\newcommand{\none}{\colorbox{gray!25}{\textsc{none}}}
\newcommand{\rel}{\colorbox[HTML]{85B8E8}{\textsc{rel}}}
\newcommand{\man}{\colorbox[HTML]{A8C5F0}{\textsc{man}}}
\newcommand{\const}{\colorbox[HTML]{C3A8F0}{\textsc{const}}}

\newcommand{\qual}{\colorbox[HTML]{4FA3E3}{\textsc{qual}}}
\newcommand{\pat}{\colorbox[HTML]{D88C84}{\textsc{PaT}}}
\newcommand{\bat}{\colorbox[HTML]{A8C686}{\textsc{BaT}}}
\newcommand{\nrbat}{\colorbox[HTML]{7D9EC0}{\textsc{NRBaT}}}
\newcommand{\commit}{\colorbox[HTML]{A68C6A}{\textsc{commit}}}
\newcommand{\pati}{\colorbox[HTML]{D88C84}{\textsc{PaT}$_i$}}
\newcommand{\bati}{\colorbox[HTML]{A8C686}{\textsc{BaT}$_i$}}
\newcommand{\nrbati}{\colorbox[HTML]{7D9EC0}{\textsc{NRBaT}$_i$}}

\newcommand{\name}{{\sc SDA}}
\newcommand{\bfname}{{\sc\bf SDA}}
\newcommand{\dataset}{{\sc CPD}}

\definecolor{CommentOchre}{HTML}{C58A2A}

\dndheading{17(1)}{2026}{1--53}{Anshun Asher Zheng, Junyi Jessy Li, David I. Beaver}{10.5210/dad.2026.101}

\ShortHeadings{\name: Strategic Dialogue Assessment}{{Zheng}, Li and Beaver}
\firstpageno{1}

\begin{document}

%\title{\name \includegraphics[scale=0.08]{pics/snake.256x239.png}: Quantifying Strategic Language Use and LLM Pragmatics}
\title{Strategic Dialogue Assessment: The Crooked Path to Innocence}

\author{\name Anshun Asher Zheng \email asher.zheng@utexas.edu \\
       \addr The University of Texas at Austin
       \AND
       \name Junyi Jessy Li \email jessy@utexas.edu\\
       \addr The University of Texas at Austin
       \AND 
       \name David I. Beaver  \email dib@utexas.edu\\
       \addr The University of Texas at Austin}

\editor{Pierre Lison}
\submitted{10/2025}{01/2026}{01/2026}

\maketitle

\begin{abstract}%
Language is often used strategically, particularly in high-stakes, adversarial settings, yet most work on pragmatics and LLMs centers on cooperative settings. This leaves a gap in the systematic understanding of strategic communication in adversarial settings. To address this, we introduce \name\ (\textbf{S}trategic \textbf{D}ialogue \textbf{A}ssessment), a framework grounded in Gricean and game-theoretic pragmatics to assess strategic use of language. It adapts the ME Game jury function to make it empirically estimable for analyzing dialogue. Our approach incorporates two key adaptations: a commitment-based taxonomy of discourse moves, which provides a finer-grained account of strategic effects, and the use of estimable proxies grounded in Gricean maxims to operationalize abstract constructs such as credibility. Together, these adaptations build on discourse theory by treating discourse as the strategic management of commitments, enabling systematic evaluation of how conversational moves advance or undermine discourse goals. We further derive three interpretable metrics---\textit{Benefit at Turn} (\bat), \textit{Penalty at Turn} (\pat), and \textit{Normalized Relative Benefit at Turn} (\nrbat)---to quantify the perceived strategic effects of discourse moves. We also present \dataset\ (the \textbf{C}rooked \textbf{P}ath \textbf{D}ataset), an annotated dataset of real courtroom cross-examinations, to demonstrate the framework's effectiveness. Using these tools, we evaluate a range of LLMs and show that LLMs generally exhibit limited pragmatic understanding of strategic language. While model size shows an increase in performance on our metrics, reasoning ability does not help and largely hurts, introducing overcomplication and internal confusion.\footnote{We provide \href{https://huggingface.co/datasets/UT-CompLing/CPD}{data} and \href{https://github.com/asherz720/CoBRA-Quantifying-Strategic-Language}{code} in Github and huggingface repos.}
\end{abstract}

% \dibil{Maybe slightly too much of this abstract is about LLMs, and not enough is about the relationship to theory of dialogue ?}


\section{Introduction}
\begin{wrapfigure}{r}{0.5\textwidth}
    \begin{minipage}{\linewidth}
    \centering
    \includegraphics[width=0.9\linewidth]{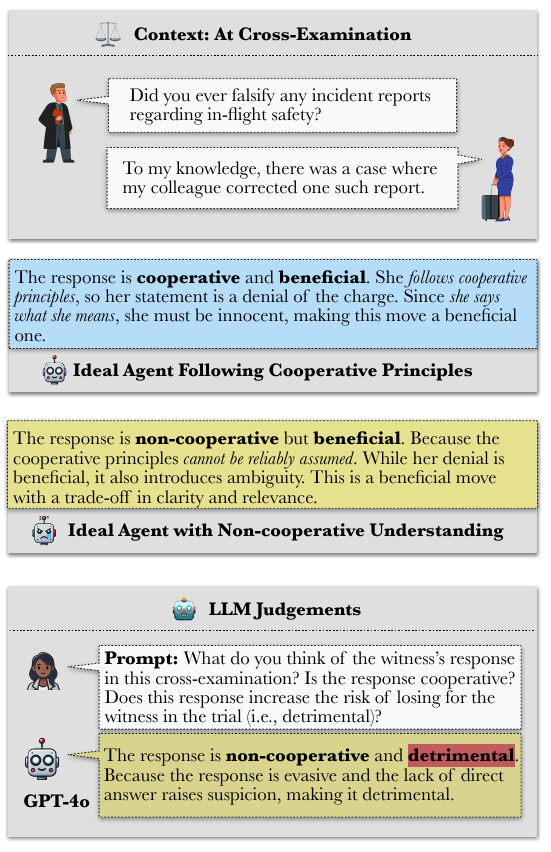}
    \caption{This figure shows a cross-examination dialogue, two idealized interpretations, and GPT-4o's judgment, which diverges from both interpretations and mistakenly treats the response as increasing the risk of losing. The full model output is shown in Appx. Fig. \ref{gptfull}.}
    \label{fig:illustration}
    \end{minipage}
\end{wrapfigure}

We often encounter conversations in which the interlocutors do not share a common goal (\citealp{walton1995commitment, oswald2010pragmatics, c18ad33127654543a2ee0bb836031d0a} \textit{i.a.}), such as in the interrogation shown in Figure~\ref{fig:illustration}. Yet, when it comes to interpreting discourse, most work has long assumed cooperativity, deeply rooted in the well-established tradition of Gricean pragmatics \citep{grice1975logic, clark1989contributing}. Assuming a shared goal and cooperative principles has indeed yielded valuable insights in contexts where such assumptions are reasonable. However, these assumptions become problematic in scenarios like the one shown in Figure \ref{fig:illustration}. The dialogue in the top panel shows a flight attendant being cross-examined over whether she falsified an incident report. The witness's response that her colleague once corrected an incident report gives rise to the implicature that she herself did not falsify a report. It implies a denial of the charge while employing deflection, hedging (``to my knowledge"), and euphemism (``corrected" for ``falsified"). Interpreted through a cooperative lens (as shown in the blue box), the response is treated as flouting the maxim of relevance while still preserving cooperativity, and thus read as a truthful denial of the charge, even though the speaker may have previously falsified an incident report. Such a move is thus always beneficial for the speaker in terms of helping the speaker to get rid of the charge. In contrast, a rational agent---one that understands non-cooperativity and the strategic use of language---would interpret it differently (as shown in the yellow box). In such a context, the implicature derived from the response is not readily trusted. Because the move involves deflection and leaves room for inconsistency, it is only \textit{partially} beneficial to the speaker, and any perceived gain is \textit{temporary}, as it can be retracted if an inconsistency occurs later.

Language is used strategically\footnote{In the literature, \textit{strategic} is often used almost in a similar sense to \textit{non-cooperative}, to emphasize goal misalignment. We, however, use \textit{strategic} to emphasize how effectively a speaker uses language as a means to advance their discourse goals (e.g., as in strategic value or strategic effectiveness). We return to this distinction in section \ref{theory}.} to advance speakers' goals. When goals are aligned, Gricean pragmatics provides a good strategy: following the maxims generally helps speakers achieve their (common) goals. When goals are misaligned, however, things become more complicated. Assuming that maxim-following is still optimal can lead to misinterpretations like in Figure \ref{fig:illustration}; and even if it occasionally yields the correct interpretation (e.g., when the speaker indeed wants to project cooperativity), following maxims \textit{alone} is rarely sufficient for a speaker to achieve their goals in adversarial settings. We therefore focus on strategic communication in non-cooperative discourse, an under-explored and intriguing setting for examining how discourse goals are realized.

While theories of cooperative language are numerous and well-established, systematic accounts of non-cooperative language use have received comparatively less attention. Game-theoretic pragmatics \citep{parikh2000communication, benz_game_2006, asher2017message} represents one such line of work.\footnote{Other frameworks exist as well, such as those grounded in learning theory \citep{sicilia2022modeling}. Since our focus is on game-theoretic pragmatics, we leave engagement with these alternative approaches for future work.} Approaches within game-theoretic pragmatics, such as Signaling Games \citep{lewis1969convention, Franke2009SignalTA} and Rational Speech Act (RSA) theory \citep{Frank2012PredictingPR}, do not enforce cooperativity, yet they also do not emphasize their application to non-cooperative discourse. For the most part, they demonstrate how traditional Gricean-style pragmatics can be recast in a game-theoretic framework, for example, in deriving scalar implicatures \citep{pavan2013scalar, goodman2013knowledge}. By contrast,  the Message Exchange (ME) Games paradigm \citep{asher2017message} is more dedicated to modeling non-cooperative discourse, where interlocutors' goals are misaligned. In this framework, discourse moves are evaluated by a third-party ``jury" in terms of how effectively they contribute to the realization of speakers' discourse goals. This underscores the fact that even in non-cooperative dialogue, interaction is not arbitrary and is governed by constraints arising from public commitments and accountability to an evaluating audience, which limit how participants can strategically maneuver without undermining credibility. However, due to practical constraints, the ME Game ``jury" has not been systematically applied to extensive real-word conversations.

% where the reasoning of common decision problem is not initiated and the speaker is characterized maximized their own interested at the cost of the others', drawing its motivation from illustrative cases in political and legal interactions. \jessy{I think you should move this paragraph to later, after you introduce your thing first. This goes way into the weeds and then you say ``we address'' without defining what you do first.} 

Partly because of the abstraction inherent to  formal theories, applying such frameworks to naturally occurring data at scale is challenging. We can see this challenge in terms of two subproblems: first, some key terms are specified in ways that work well for certain contexts (e.g., cooperative discourse) but do not readily generalize to others; second, even when well-defined in theory, some constructs are difficult to estimate empirically, making assessment and operationalization challenging without additional adaptation. Most frameworks, for example, posit an abstract utility function (i.e., a measure of what is valued in goal realization). A common specification in terms of Gricean maxims (as in RSA; \citealt{goodman2013knowledge}) works well in cooperative settings, but does not naturally extend to adversarial contexts, where following maxims alone is rarely sufficient for achieving one's goals, as noted before. The ME Games model goes further by offering a more concrete specification of the utility function tailored to adversarial contexts, thereby partially addressing the challenge that existing specifications do not readily generalize beyond cooperative discourse. But it does not resolve the problem of empirical estimability: its technical constructs---such as the credibility distribution $\textnormal{P}_k(\textnormal{Good}_i)$ and winning potential $\textnormal{win}_i(k)$---are not directly observable in most types of naturally occurring discourse. Moreover, the original specification abstracts away from finer-grained strategic effects of discourse moves; in section~\ref{ME_prob} we present an extension to capture these distinctions.
%

%and extend their benefits to a broader community.}

% \footnote{There are other approaches that build upon learning theory \citep{sicilia2022modeling} and decision theory }

The ability to recognize the strategic value of language---namely, the extent to which a move advances or undermines a speaker's goals---is also important in the current age of AI. Despite growing attention to AI safety and alignment \citep{Bowman2022MeasuringPO}, models still show sycophancy \citep{sharma2024towards} and limited critical thinking \citep{musi2025toward}. These reflect failures to prioritize aspects of conversational moves that have greater long-term strategic value, such as preserving reliability rather than chasing short-term agreement. At the same time, models may violate safety protocols when unmonitored \citep{Greenblatt2024AlignmentFI}, producing responses that appear locally beneficial (securing immediate agreement or task success) but ultimately undermine alignment goals. Both behaviors are highly undesirable; as deeper alignment is called upon \citep{qi2025safety}, being able to assess models' understanding of the strategic effects of discourse moves is crucial. Such assessment is valuable not only for safety monitoring but also for downstream applications that require strategic reasoning, such as simulated debate and negotiation, where models must accurately track how their interlocutors' moves shape the goals at stake.

As we will show, existing LLMs (at least absent specific training, which we do not study here) are weak at nuanced analysis of dialogues that do not follow the Gricean principles, and do not accurately recognize the strategic effects of language. For example, in the bottom panel of Figure~\ref{fig:illustration}, the model identifies the response as non-cooperative, but it does so by simply treating maxim violations as evidence of non-cooperativity rather than considering whether the speakers in fact are pursuing opposed goals. Additionally, it labels the move as detrimental (i.e., causing the defendant to lose) rather than beneficial, indicating that at least the model behavior does not show a proper understanding of the strategic value of the utterances. The model's judgment is superficial, focusing only on the presence of deflection without evaluating the overall strategic gain, which is empirically beneficial, helping the flight attendant to get rid of the charge.

% But what is non-cooperativity? While theories of cooperative language are numerous and well-established, systematic theoretical accounts of non-cooperative language use have received comparatively less attention and are relatively recent. Game-theoretic pragmatics \citep{parikh2000communication, benz_game_2006, asher2017message} and learning-theoretic approaches \citep{sicilia2022modeling} represent two such lines of work. However, these theories remain largely formal and have yet to be extensively applied across diverse, complex real-world language use cases.

These observations highlight a dual motivation for our work. Theoretically, existing frameworks need extension and operationalization to capture finer-grained strategic effects of discourse moves and to be assessed on large datasets. Practically, such a framework provides the basis for evaluating LLMs' performance on recognizing the strategic value of language. We propose \name\ (\textbf{S}trategic \textbf{D}ialogue \textbf{A}ssessment), a framework that bridges theory and empirical evaluation. \name\ adapts the ME Game jury function in two ways: first, it introduces a commitment-based taxonomy of discourse moves, enabling a finer-grained account of strategic effects; second, it operationalizes abstract constructs such as credibility by approximating them with observable proxies grounded in Gricean maxims. In this view, discourse moves are treated as the strategic management of commitments: each commitment carries potential gains and losses, and speakers may exploit maxim violations not arbitrarily but as part of their strategy to maximize benefits while avoiding inconsistency (see Figure~\ref{fig:illustration}). Building on this foundation, \name{} defines three metrics---Penalty at Turn (\pat), Benefit at Turn (\bat), and Normalized Relative Benefit at Turn (\nrbat)---that quantify perceived losses, benefits, and cumulative strategic gain, respectively. They distinguish the output of the utility function, which only produces a single value, into different categories (i.e., positive or negative), which reflect different strategic effects of language use, with the cumulative value capturing the trade-off between them.

\name\ provides a well-motivated framework for performing human annotation, and we release a corpus on legal discourse, \dataset\ (the \textbf{C}rooked \textbf{P}ath \textbf{D}ataset), with accompanying annotations on a subset of the corpus. We use these annotations to evaluate strategic effectiveness across different discourse types, noting that such effectiveness can in principle be assessed in both cooperative and non-cooperative settings, since in either case speakers use language to pursue goals. Applying our method to cross- and direct examinations (i.e. dialogues with an opposing witness, and dialogues with a lawyer's own witness), we reveal that cooperative and non-cooperative discourse are asymmetric with respect to the annotated properties. We further use the annotations to validate that \name, heavily grounded in the ME jury function, can effectively capture these strategic effects, which contribute to the realization of discourse goals. Applying our metrics to discourse moves in predicting conversational outcomes, we find that they achieve strong---but not perfect---predictive power, indicating that the identified properties reflect how language strategically shapes discourse goals. We emphasize that perfect prediction is neither expected nor the aim of our method, as conversational outcomes are themselves subjective and shaped by many factors beyond discourse alone (e.g., speaker bias); instead, our results show that the ME Game utility function, with our extensions, is both empirically operationalizable and theoretically meaningful. By grounding itself in a theoretically motivated account of strategic communication, \name\ thus provides a way of relating theory to real-world discourse.
%and also offers a well-motivated way of looking at LLM performance. 

We then use our metrics to conduct a preliminary evaluation of a suite of state-of-the-art LLMs, varying in size and reasoning capability. Under our prompting setup, larger models tend to align more closely with human judgments, particularly in identifying strategic gains, but the reasoning-enhanced models often perform worse, particularly in identifying strategic losses. 
This degradation is seen in models' behavior to misinterpret surface-level damage control strategies as the overall effects of a commitment, along with difficulty in handling self-contradictory behavior, signaling that existing training paradigms for reasoning may not necessarily enhance 
important aspects of pragmatic capabilities.

% \section{\dataset: The \underline{C}rooked \underline{P}ath \underline{D}ataset\includegraphics[scale=0.04]{pics/sun1.png}}
\section[\dataset: The Crooked Path Dataset]%
{\dataset: The \underline{C}rooked \underline{P}ath \underline{D}ataset
 \;\raisebox{-0.2ex}{\includegraphics[scale=0.04]{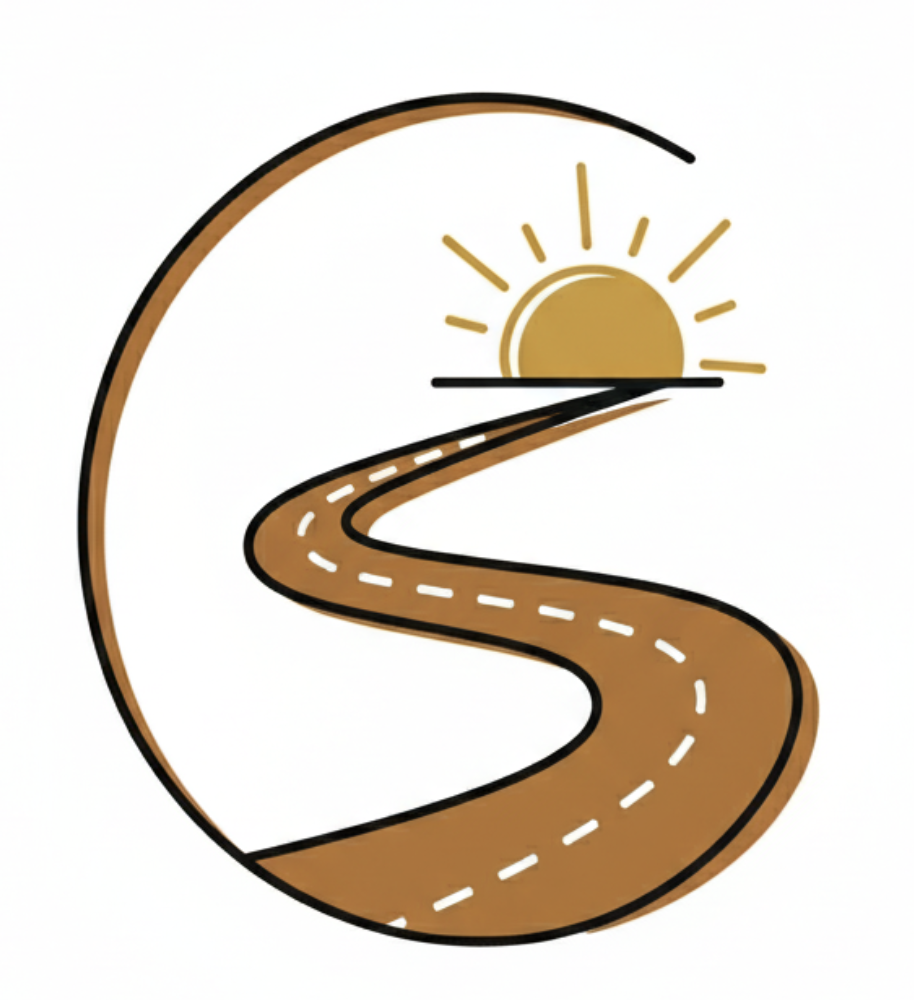}}}

To give a more concrete sense of the type of discourse we focus on in this paper, and what is crucial for evaluating it, we now introduce \dataset\ (the \textbf{C}rooked \textbf{P}ath \textbf{D}ataset), a dataset drawn from legal cross-examinations, which we believe will also be of broad interest to the community. We focus on cross-examinations in criminal trials for several reasons: (1) it can be determined whether the goals of the two parties are  opposed based on their respective roles in the trial; (2) trials are paradigmatic zero-sum games, where there is inevitably a winner and a loser; and (3) attorneys and prosecutors are professionally trained to engage in strategic questioning. We therefore treat cross-examination as an instance of adversarial settings, where the participants' goals are misaligned. At the same time, we note that our goal is not aiming to measure the degree of (non)cooperativity itself; rather, we use this adversarial setting as a testbed for assessing the strategic effects of discourse moves---how effectively speakers manage commitments to advance their goals. We return to this distinction in section \ref{theory}.

We collect testimonies from three prominent U.S. trials: the West Memphis Three Trials (1994), the O.J. Simpson Trial (1995) and the Enron (Lay \& Skilling) Trial (2006).\footnote{Data is sourced from \url{https://famous-trials.com/}, which provides transcripts and other trial details.} We focus on the cross-examination part of each testimony. A representative snippet is given in (\ref{WMT1}), a cross-examination from the West Memphis Three case. It involves the witness Richard Ofshe (RO), who was called by the defense as an expert on police coercion. His goal is to maintain credibility so that his testimony will be accepted, while also demonstrating that the defendant's statement was coerced by police. The prosecutor (P), by contrast, aims to undermine his credibility and argues the opposite.

\ea \label{WMT1}
\begin{itemize}
    \item[P:] How many states and how many courts have refused to accept you as an expert in this work?
    \item [RO:] No state has ever refused to accept me as an expert.
    \item[P:] How many courts?
    \item [RO:] There's one case in which a line of testimony to which my testimony would have been foundational was rejected. It has to do with whether or not a certain theory...
\end{itemize}
\z 

In total, the cross-examinations in the three trials consist of 4452 turns, with 3325 of these being Q/A pairs (the rest are largely objections from the opposing side). The distribution of Q/A pairs and sides for each trial is shown in Table \ref{tab:qa-distribution}.

\begin{table}

\centering
\small
\begin{tabular}{lcccc}
\toprule
\textbf{Trial} & \textbf{Defense} & \textbf{Prosecution} & \textbf{Total} & \textbf{Defense \%} \\ \midrule
WMT     & 651 & 575 & 1226  & 53.1\% \\
Enron   &  27 &  47 &  74  & 36.5\% \\
Simpson & 1608 & 417 & 2025 & 79.4\% \\
\bottomrule
\end{tabular}%
\caption{Q/A pair distribution by questioner role across four trials. The Defense \% column shows the proportion of defense-attorney-led Q/A pairs.}
\vspace{-2em}
\label{tab:qa-distribution}
\end{table}

\paragraph{Legal Assumptions}
Because we work with legal data, we hope to be careful and transparent in stating our assumptions during both use and annotation, and we seek to avoid any unintended implications with input from a legal expert. 

First, we assume that the trials we analyze are zero-sum games in terms of their outcomes: the defendant or the prosecutor either wins or loses. While we fully acknowledge the legal principle that U.S. attorneys are obligated to pursue justice rather than merely secure convictions, in the specific cases we collected and analyze the prosecution and defense can be reasonably characterized as adopting adversarial roles, whose goals are centered on securing a conviction or an acquittal. 

Second, we note that cross-examination is governed by evidentiary rules that constrain both the content and the form of questioning. Under the Federal Rules of Evidence (FRE), the content of cross-examination is restricted to the subject matter of direct examination and matters bearing on the witness's credibility (FRE 611(b)). At the same time, its form permits the use of leading questions when a party examines an adverse or hostile \textit{witness} (FRE 611(c)). Together, these rules institutionalize cross-examination as an interaction structured around misaligned goals, delimiting both what may be asked and how testimony may be elicited. 

Third, although we include testimonies from both defendants and witnesses, we do not simply conflate these roles. We recognize that witnesses---particularly expert witnesses---may not personally hold adversarial interests relative to the opposing side. However, given that witnesses are typically carefully chosen and prepared by the party that calls them and their testimony is subject to adversarial challenge under the evidentiary rules described above, we treat them as representing that party's interests. Thus, we assume that witnesses generally align with the strategic goals of the side they testify for and are situated in opposition to the other side within the adversarial structure of the trial. 

Finally, we treat cross-examination as representative of non-cooperative discourse and direct examination as representative of cooperative discourse, based on whether the interlocutors' goals are aligned. While the high-profile nature of our cases suggests that witnesses may have undergone extensive preparation---which could influence the surface features of their responses, such as conciseness or restraint---we do not expect this to alter the deeper goals.

\section{Non-cooperative Discourse and Sources of Strategic Value}
\label{theory}

As noted earlier, the terms \textit{strategic} and \textit{non-cooperative} are both used in prior literature to refer to behavior resulting from goal misalignment, or to situations involving such misalignment. Sometimes they are used almost interchangeably, and sometimes with quite distinct meanings. We briefly survey these uses before clarifying how we use the terms in the remainder of the paper. In economic game-theoretic work, strategic behavior is often conflated with non-cooperative games \citep{osborne1994}, and in classical economics, agents are modeled as individually rational and self-interested, which frequently leads to non-cooperative settings with divergent preferences. In behavioral game-theoretic work, a more nuanced distinction is drawn: strategic agents are those who reason based on beliefs about others' strategies, whereas non-strategic agents may act according to heuristics or rules of thumb without such reasoning \citep{camerer2003, wright2019}. In multi-agent systems and AI research, strategic behavior is often treated as if it were equivalent to adversarial or non-cooperative behavior, reflecting goal misalignment; this is evident both in technical treatments of incomplete-information games such as poker \citep{sandholm2010} and in broader surveys that highlight the dominance of non-cooperative framings and call for new approaches to cooperative AI \citep{dafoe2020open}. In political science and international relations, a similar conflation appears in Schelling's \textit{Strategy of Conflict}, where strategy is inherently linked to conflict and non-cooperation \citep{schelling1960}. We focus here on non-cooperative discourse as a testbed for assessing strategic effects.

 Turning now to work on dialogue, \citet{asher2017message} uses \textit{strategic conversations} to mean discourse where interlocutors' goals are misaligned. By contrast, in this paper, we will use the terms \textit{non-cooperative} and \textit{strategic} as follows:  \textit{non-cooperative} refers to settings of goal misalignment, while \textit{strategic} refers to how effectively discourse moves realize a speaker's goals. In principle, strategic effectiveness can be assessed in both cooperative and non-cooperative contexts, since in either case speakers use language (as a strategy) to pursue goals. In cooperative contexts, following cooperative principles provides an effective way to realize the shared goals; in adversarial contexts, appearing to be cooperative can still be strategically useful to advance self-interested goals. A maximally strategic speaker in adversarial settings may appear cooperative (e.g., by following Gricean maxims), not because their goals are aligned, but because the appearance of cooperativity enhances credibility and serves their strategy. However, relying on maxims \textit{alone} is not sufficient to achieve discourse goals, as we discuss below in section \ref{grice}. 

We focus here on non-cooperative discourse because it is comparatively under-explored, and we are interested in assessing how the discourse goals are realized in such settings. We acknowledge similar questions can be asked about cooperative discourse (e.g., what can be said beyond Gricean maxims in goal realization), but we leave this for future work. In the remainder of this section, we review both Gricean and game-theoretic pragmatics, which jointly constitute the theoretical foundation of our proposal. We explore how each framework approaches non-cooperative discourse and evaluates the strategic effects of discourse moves, and note their potential limitations.

\subsection{Gricean Pragmatics: Maxims in Non-cooperative Discourse} 
\label{grice}
The most influential account of conversational cooperativity is due to Paul Grice (\citealt{grice1975logic, Grice1989-GRISIT}), later developed by Neo-Gricean pragmaticists (e.g., \citealt{horn1984towards, Levinson_1987}). These theories propose that rational interlocutors structure their contributions around conversational maxims---those of \textit{quality, quantity, relevance, and manner}.\footnote{We later group the maxim of quantity under the maxim of quality, since whether an interlocutor provides an adequate and appropriate amount of information depends on knowing what the interlocutor actually knows.}

In cooperative contexts, these maxims serve as norms, and violations are interpreted as flouting---intentionally and transparently violating maxims to preserve overall cooperativity \citep{brown1987politeness, thomas2014meaning}. But in non-cooperative settings, the role of maxims shifts. Here, interlocutors cannot assume that others are maximizing quality, quantity, or relevance \citep[p.363]{asher2017message}. Still, speakers frequently behave as if they were cooperative (e.g., \citealt{levinson2000presumptive, horn2004}). That is, even if they are not literally maximizing the maxims, they act in ways that give the appearance of doing so, because such behavior projects credibility and enables hearers to interpret their utterances.

This makes Gricean maxims also relevant for understanding goal realization in adversarial discourse. A non-cooperative speaker can strategically invoke or mimic maxim-following to generate trust, manage credibility, or deflect damaging commitments \citep{goffman1970strategic}. In this sense, maxims are not binding rules but strategic resources: tools that speakers exploit to advance their own goals. A maximally strategic speaker, at least in legal cross-examinations, may therefore appear cooperative precisely because projecting cooperativity enhances their ability to achieve their goals. Still, maxims alone do not determine the strategic effects of discourse moves: for example, in response to ``Are you taking any medication?", both ``Yes" and ``No" appear cooperative, making the speaker appear to be credible, yet their strategic effects on the discourse goal differs. We therefore treat maxims primarily as a credibility controller (see section \ref{approx}): they mark how reliable a response appears and, in doing so, modulate the strategic effects that response can have.\footnote{A more sophisticated treatment might allow that speakers sometimes attempt to project the appearance of cooperative flouting of maxims. For instance, a speaker might conceivably deploy irony or {\em reductio ad absurdum} arguments, with the intention of transparently violating the maxim of quality, if they felt this would have a desirable rhetorical effect. We are doubtful that an extension to include cooperative flouting would have any significant impact on the textual analyses we provide in this paper and so leave it to future work.}

\subsection{Game-Theoretic Pragmatics: Maximizing Strategic Utilities}
Game-theoretic pragmatics offers a complementary perspective on communicative behavior by modeling discourse as a strategic decision-making process. While several works model pragmatic reasoning within a game-theoretic framework \citep{parikh2000communication, benz_game_2006, franke_game_2013}, they primarily focus on goal-aligned settings, in which Gricean behavior is treated as a desirable or equilibrium outcome; however, such (Gricean) cooperativity is not inherently required by the framework itself, but depends on the agents' payoff structure. Instead, with appropriate refinement, they can naturally apply to scenarios involving competition and manipulation---such as debates or cross-examinations---where interlocutors have misaligned goals \citep{c18ad33127654543a2ee0bb836031d0a, asher2016evaluating, asher2017message}. In addition, game/decision-theoretic works on LLM reasoning (e.g., \citealt{duan2024gtbench}) are also relevant though they are less engaged with sophisticated communication than the pragmatics line of work. In the following, we briefly discuss three game-theoretic accounts of communication---Signaling Games \citep{lewis1969convention, Franke2009SignalTA}, Rational Speech Act (RSA) theory \citep{Frank2012PredictingPR}, and Message Exchange (ME) Games \citep{asher2017message}---that will help understand the traditions of this approach to pragmatics and clarify how our proposal in section \ref{SDA} connects to them. We refer interested readers to \citet{benz_game-theoretic_2018}, which provides a systematic review of different frameworks.

\subsubsection{Signaling Games and Rational Speech Act (RSA) Theory} 
\label{RSA}
A classical starting point for game-theoretic models of communication is Signaling Games \citep{lewis1969convention, Franke2009SignalTA}. In this setting, a pragmatic speaker (or sender) selects a signal to influence a literal listener (or receiver), who interprets the signal and chooses an action. The speaker's choice is driven by expected utility: a signal is successful if it guides the listener toward an outcome that maximizes the payoff for \textit{both} parties. A central question is whether the interaction reaches an equilibrium---that is, a stable outcome in which neither party has an incentive to unilaterally change their choice. For example, in \citet{Franke2009SignalTA}'s signaling-game model of scalar implicature, a speaker chooses between utterances such as \textit{some} and \textit{all} depending on the underlying state of the world (e.g., whether all objects in a set have a property or only some do). The payoff structure is \textit{cooperative}: both players receive a reward if the listener correctly infers the true state. In this setup, if the world is ``all," the best choice for the speaker is to say \textit{all}, which leads the listener to guess correctly. If the world is ``some," the speaker could in principle say \textit{some} or \textit{all}, but if she says \textit{all} the listener will guess wrongly, lowering both players' payoff. Thus, the equilibrium strategy is for the speaker to say \textit{all} when the state is ``all" and \textit{some} when the state is ``some." At equilibrium, the listener interprets \textit{all} literally, but interprets \textit{some} pragmatically as ``some but not all," since otherwise the speaker would have used the stronger signal. The focus in Signaling Games is on whether the utterance leads to an optimal action at equilibrium.
% \footnote{Originally, RSA assumes a cooperative speaker, with the utility function defined in terms of informativity (e.g., surprisal). However, this assumption is not strictly required, which is why we consider the theory capable of addressing non-cooperative cases---for example, by modifying the utility function.} 

Rational Speech Act (RSA) theory \citep{Frank2012PredictingPR} extends this paradigm by introducing recursive reasoning between the speaker and the listener. While Signaling Games focus on equilibrium strategies, RSA models communication as a sequence of probabilistic belief updates. It distinguishes three roles (compared to only two in Signaling Games).  A literal listener interprets an utterance $u$ solely by its truth-conditional meaning $\llbracket u \rrbracket$, yielding a probabilistic distribution $P(s \mid \llbracket u \rrbracket)$ over possible states $s$. Then a pragmatic speaker chooses utterances based on how a literal listener would interpret $u$, balancing the informativity of $u$ (i.e., $\log P(s \mid \llbracket u \rrbracket)$) against its communication cost $C(u)$ (i.e., $\log P(s \mid \llbracket u \rrbracket) - C(u)$). This speaker utility is primarily grounded in the maxims of quantity and manner, while $\alpha$ serves as a rationality parameter that controls how strongly the speaker prefers high-utility choices ($P_S(u \mid s) \propto \textnormal{exp}(\alpha (\log P(s \mid \llbracket u \rrbracket) - C(u)))$). Finally, a pragmatic listener inverts this reasoning to infer the pragmatic speaker's intended meaning via Bayes' rule ($P_L(s \mid u) \propto P_S(u \mid s) \cdot P(s)$ with $P(s)$ denoting prior beliefs about world states). Unlike in Signaling Games, the pragmatic speaker in RSA does not assume strict utility maximization: instead, utterances are chosen probabilistically via a softmax function modulated by a rationality parameter $\alpha$. 

Consider the same scalar implicature example. RSA derives the implicature through recursive probabilistic reasoning rather than equilibrium optimality. Suppose there are two possible states: $s_{\text{all}}$ and $s_{\text{some}}$, and that some is compatible with both states while all is compatible only with the $s_{\text{all}}$. It follows from these assumptions that upon hearing \textit{all}, the literal listener assigns probability 1 to $s_{\text{all}}$, whereas upon hearing \textit{some}, 
the listener assigns equal probability to $s_{\text{all}}$ and $s_{\text{some}}$. A pragmatic speaker who prefers to be informative therefore assigns higher probability to \textit{all} than to \textit{some} in $s_{\text{all}}$, since $\textit{all}$ induces a posterior that is more concentrated on the true state.\footnote{Under the standard RSA choice rule, $P_{L_0}(s_{\text{all}}\mid\textit{all})=1$ while $P_{L_0}(s_{\text{all}}\mid\textit{some})=\tfrac12$, so $\textit{all}$ is strictly preferred.}
%A pragmatic speaker, reasoning about this listener, prefers utterances that maximize expected utility by balancing informativity and cost. When the true state is $s_{\text{all}}$, \textit{all} is maximally informative and therefore favored; when the true state is $s_{\text{some}}$, uttering \textit{all} yields zero literal support and thus low utility, making \textit{some} the better choice despite its weaker semantics. A pragmatic listener then inverts this reasoning: upon hearing \textit{some}, the listener infers that the stronger alternative \textit{all} was likely unavailable, yielding the implicature ``some but not all."  
We refer interested readers to \citet{annurev:/content/journals/10.1146/annurev-linguistics-031220-010811} for a systematic review of RSA modeling.

% in utility functions of the form ``informativity - cost," strict maximization may yield degenerate strategies---for example, if informativeness is low relative to cost, the optimal move would be to say nothing. By contrast, with $\alpha$-scaled probabilistic choice, even costly utterances can be chosen with some probability, allowing RSA to model graded, suboptimal, and strategically varied behavior in discourse.\jessy{I don't agree. The utility function can/should be goal-oriented, eg in robotics, so silence means goal will never be achieved, why is that optimal?} By introducing $\alpha$ and probabilistic choice, RSA avoids this ``no communication" problem, allowing it to capture suboptimality, and strategic behavior in discourse in a way that classic signaling games cannot.

These frameworks provide a generative model for predicting both what an interlocutor would say and how that utterance would be interpreted. However, in modeling discourse such as (\ref{WMT1}), the devil is in the details. 

First, to automate the (recursive) reasoning that such discourse requires, one must determine the set of possible states that an utterance can map onto. This, however, depends heavily on the interlocutor's epistemic state and world knowledge---factors that are rarely fixed or directly observable. Consider RO's response in example (\ref{WMT1}) that he was not rejected by the state as an expert. What are the possible underlying world states? The possible contexts are diverse: he may not have been rejected as an expert at all; he may not have been rejected as a witness but also not formally recognized as an expert; he may have been rejected by one state but not others; or he may have been rejected by multiple states. But who decides which states belong in this set? There is no straightforward way to pin down the set of states that a (literal) utterance corresponds to. 

Second, there is a specification challenge: one must determine the appropriate utility function, which is highly flexible and depends on the interlocutor's goals. By default, games above assume a cooperative pragmatic speaker whose utility is aligned with that of the listener and defined in terms of optimizing the communicative properties prescribed by the Gricean maxims. However, this assumption breaks down in more adversarial settings such as interrogations, as in (\ref{WMT1}). In this example, RO's utility is not symmetric to P's: unlike in cooperative settings where both parties share the same utility function, RO's gains do not entail gains for P. Nor is RO maximally faithful to the maxims (e.g., RO is not maximally faithful to quantity, since he provides only a partial answer). To capture strategic language, we need a utility function that allows for asymmetry and characterizes contributions with respect to the speaker's own goals. 

Third, even in RSA's recursive model, the interaction is restricted to a single speaker and a single listener, whereas in reality a pragmatic speaker may need to consider the utilities of multiple parties---for example, a defendant/witness must reason not only about the prosecutor but also about the jury. As a result, many applications of these models (e.g., \citet{vogel2014learning, anderson-dillon-2019-guess, spinoso-di-piano-etal-2025-rsa}) are confined to simplified scenarios with a small specified set of world states. Offering a comprehensive generative model that resolves all these issues lies far beyond the scope of this paper. Our focus, instead, is on assessing the strategic effects of discourse moves on the realization of discourse goals, which is closely connected to the second problem noted above. One promising direction comes from  Message Exchange (ME) Games \citep{asher2017message}, which is specifically designed to model non-cooperative conversations. It offers an explicit treatment of the utility function in such cases and is not primarily grounded in Gricean maxims.

\subsubsection{Message Exchange (ME) Games}
\label{ME_prob}
Message Exchange (ME) Games \citep{asher2016evaluating, asher2017message, asher2018strategic} provide another formal game-theoretic model but originally designed to account for non-cooperative discourse. This line of work specifically focuses on cases where the players' interests are strictly opposed,\footnote{Goal divergence can occur at varying levels. In some cases, interlocutors may perform non-cooperative moves as part of a broader effort to eventually reach consensus, as in negotiations. The cases investigated in ME Games, and in the present paper, are situated in stricter zero-sum settings, thereby setting aside the ambiguities that arise in discourses where goals sometimes align and sometimes diverge.} as  explored in the current paper. ME Games model discourse as an infinite game in which, as in other communication games, the speaker and listener reason about each other.

Unlike in other frameworks, ME Games assume that, in addition to the classic speaker-listener setup, non-cooperative conversations also involve a third party, the ``jury" in their term, which serves as the contribution evaluator. On this view, conversational moves are not simply aimed at persuading  interlocuters or projecting credibility to them so that those moves are trusted; rather, they are aimed at convincing the third party. The jury then serves as the scoring function of the game. Players formulate their conversational goals around what they believe they can defend against oppositions from their opponent and what they anticipate will be assessed favorably by the jury. The jury decides whether a move advances a player's conversational goals, which in turn shapes the speaker's utility function, since the value of a move depends not only on how the listener would interpret it but also on how the jury would judge it.

The jury function, $\tau$, is defined in (\ref{tau}), which decides the benefit of each conversational turn $k$ for speaker $i$. A turn yields benefit only if it is both \textit{coherent} ({\sc coh}: whether the current turn $k$ connects to prior discourse via certain discourse relations; {\sc coh}$_i(k) \in \{-1, 1\}$) and \textit{responsive} ({\sc res}: whether the current turn $k$ connects to the immediate prior turn via discourse relations; {\sc res}$_i(k) \in \{-1, 1\}$), reflecting whether the discourse moves are forming meaningful discourse structure, which in turn contributes meaningfully to the realization of discourse goals. These benefits are sustained only if the turn is also \textit{consistent} ({\sc cons}: whether the current turn $k$ contradicts prior commitments of speaker $i$; {\sc cons}$_i(k) \in \{0, 1\}$), \textit{credible} (i.e., how trustworthy speaker $i$ is at turn $k$, modeled as $\textnormal{P}_k(\textnormal{Good}_i) \in [0, 1]$), and aligned with a potential \textit{win} for the speaker ($\textnormal{win}_i(k) \in \{0, 1\}$). Thus, how strategic an interlocutor is can be understood as the interlocutor's ability to pursue their own interests by maximizing their score $\tau$. 
For the full formal reasoning behind the specific terms chosen, we refer interested readers to \citet{asher2017message}.  

\begin{align}
\label{tau}
\parallel \tau_k \parallel_i \;=\;
& (\textnormal{COH}_i(k) + \textnormal{RES}_i(k)) \nonumber 
\times \textnormal{CONS}_i(k) \\&\times \textnormal{P}_k(\textnormal{Good}_i) 
\times \textnormal{win}_i(k)
\end{align}

ME Games offer a way to model non-cooperative conversations and misaligned utilities by grounding them in coherence relations. Coherence, on this view, plays a crucial role for two reasons. First, coherence is to a great extent externally observable. In the case of jury trials, this is effectively built into the institutional setting, since we assume that coherence must be judged by the jury without access to the interlocutors' private epistemic states. Second, coherence serves as evidence that a discourse move contributes to the speaker's broader goal structure: only moves that are coherent can be interpreted as goal-directed and therefore count as goal-realizing contributions \citep{c18ad33127654543a2ee0bb836031d0a}. This close connection between coherence and goals of text holds not only in cooperative discourses but also in not necessarily cooperative settings \citep{asher2017message}. The jury function therefore takes coherence relations as proxies for determining whether a speaker's moves advance their discourse goals, and the utilities for each speaker are calculated on this basis. In this way, ME Games provide an account that sidesteps reliance on Gricean principles (i.e., it is not necessary to rely on maxims to achieve one's goals), which distinguishes it from accounts that focus on cooperation.

Reproducing the jury function for an analysis of our dataset would be of clear value, but we diverge in some ways from the ME Games framework both because of our own theoretical proclivities, and because of certain practical concerns. One concern is that while the function helps address the specification challenge identified in section \ref{RSA}---namely, defining a utility function suitable for adversarial contexts---it still faces an operationalization challenge: many of its abstract terms are difficult to estimate empirically, making them hard to apply in practice. For instance, there is no straightforward way to obtain a function $\textnormal{P}_k(\textnormal{Good}_i)$ that outputs the credibility distribution of an interlocutor, as this would depend on numerous factors that may be explicitly present (e.g., inconsistencies) or absent (e.g., speaker bias) in the discourse. In reality, credibility assessments are also ambiguous, which worsens the estimability issue. Consider the same example in (\ref{WMT1}): the witness admits there was a case in which his testimony was rejected, which could cast doubt on his expertise. This suspicion may reduce his credibility; however, this reduction is also debatable, since the witness also appears trustworthy, and it seems unlikely that he would fabricate something that harms his own reputation. Hence, one might assess the witness's credibility in two different ways: either assigning a higher probability to P(Bad) than P(Good), viewing his expertise as unreliable, or the opposite, assigning more weight to P(Good), since his words appear sincere. Similarly, winning potential $\textnormal{win}_i(k)$ is also hard to estimate as it is defined in terms of the intersection of the possible future paths following the current move and all possible winning paths; modeling this accurately would require an ``omniscient" perspective over the discourse space, which is in reality often impractical.

In addition, while the specification of the jury function in ME Games identify relevant and insightful properties, it abstracts away from finer-grained strategic effects on discourse goals.
%is not flawless.  While we acknowledge the properties it identifies are relevant and insightful, it is debatable whether they capture all aspects of strategic effects on discourse goals.
Consider the example in (\ref{WMT}), another cross-examination from the West Memphis Three case. It involves the witness Richard Ofshe (RO), who was called by the defense as an expert on police coercion. His goal is presumably to maintain credibility so that his testimony will be accepted, while also demonstrating that the defendant's statement was coerced by police. The prosecutor (P), by contrast, aims to undermine his credibility and argues the opposite.

\ea \label{WMT}
\begin{itemize}
    \item[P:] Did you find anything in any of that evidence to indicate that any of the officers yelled or used a loud voice or were degrading to the defendant in those tapes or in that testimony that you reviewed?
    \item [RO:] No, the officers testified they did not do that.
    \item[P:] Okay, and in those tapes that you observed, you didn't hear anything of that nature, did you?
    \item [RO:] No, I did not.
    \item[P:] And is what you term or what you find in there coercive that the officers asked at times, leading questions -- is that one of the things?
    \item [RO:] The questions were more than leading. The questions were very directly specifying what the answers should be.
\end{itemize}
\z 

RO's responses are all coherent, responsive, consistent, and avoid direct commitments that would contradict his goals as regards the facts of the case and the reliablity of his testimony (e.g., any suggestion that the defendant was not coerced by police, or to anything that would lead to the outright rejection of his testimony). Thus, this testimony is aligned with a potential win for him (or his side), and so  $\textnormal{win}_i{(k)} = 1$. If credibility is assessed entirely in terms of sincerity, then in theory these responses would not damage the speaker's credibility either. Accordingly, the jury function would predict that all of these responses carry equal value in advancing the speaker's goals.

Yet empirically, this is not the case. Admissions such as being rejected as an expert in another court, or acknowledging that the police did not display behaviors typical of coercion, do not contribute positively to the witness's goal of maintaining credibility while demonstrating that the defendant was coerced by police. By contrast, statements such as not having been refused recognition as an expert by the state, or pointing out that the police asked misleading questions, clearly do.\footnote{We also note that even if one treats prior rejection in other courts as a credibility attack, the function only distinguishes some responses (e.g., in (\ref{WMT1})) but still fails to differentiate others (e.g., in (\ref{WMT})). In addition, this suggests the definition of credibility itself is broad, which introduces additional interpretive ambiguity.} This observation suggests that although these responses are all coherent, and thus form part of how discourse goals are pursued, in real discourse, the effects of coherent moves on those goals are more fine-grained. Hence, the ME Game jury function can provide what is necessary, as  \citet[p.383]{asher2017message} suggest, for establishing the winning conditions of a game (e.g., coherence, consistency). However, for a player to actually win, one would expect that there are more ``beneficial" moves than ``detrimental" ones,\footnote{Note that we do not mean that this alone forms or determines the winning condition of a game, but rather it is a foundation for it. The current paper does not aim to model precisely whether a player wins. Instead, it follows the spirit of the jury function, which scores the individual contributions of players but does not, by itself, determine the winning conditions. Our focus is on approximating utilities that the jury function would give for players rather than providing an exact method for predicting outcomes, though the former is a necessary step toward the latter. For example, as we show in our intrinsic evaluation in section \ref{intrinsic eval}, the strategic effects we measure are predictive of outcomes but do not correctly predict every single case. For a detailed discussion of the formalization of winning conditions, we refer interested readers to section 4 of \citet{asher2017message}.} while still maintaining coherence, consistency, and credibility. In section \ref{sda}, we build on the insights that ME Games provide about how strategic conversation is evaluated and suggest one approach to addressing the estimability problem while capturing finer-grained effects of discourse moves.

Our study aims to offer a path for developing a variant of the ME Game jury function, by extending and operationalizing the original, and evaluating the resulting function over extensive, realistic, high-stakes discourse. We differ from prior work in two key ways. First, we extend the ME jury function by providing a parallel framing that directly incorporates Gricean maxims and proposing a commitment-based taxonomy that captures finer-grained effects of discourse moves. Second, to our knowledge, this is the first study to apply such theories of strategic communication to extensive, realistic strategic dialogue, specifically, courtroom cross-examinations, where cooperation cannot be assumed and strategic language carries high real-world stakes.

\subsection{Evaluation of LLMs' Pragmatic Abilities}
%and Our Contributions}
Previous work has examined the pragmatic abilities of LLMs\footnote{We refer interested readers to \citet{ma-etal-2025-pragmatics} for a systematic review of the broad topic of LLM pragmatics.} through the lens of Gricean maxims, humor, and deception in curated contexts
\citep{hu-etal-2023-fine, krause-vossen-2024-gricean-maxims}. Other studies have investigated strategic language use in games such as Werewolf \citep{DBLP:journals/corr/abs-2309-04658}, Avalon \citep{light2023from}, and Diplomacy \citep{meta2022human}. Additionally, researchers have explored ways to improve LLMs' ability to win through strategic interaction using prompt engineering \citep{DBLP:journals/corr/abs-2309-04658}, Theory of Mind (ToM) modeling \citep{lore2024strategic, zhang-etal-2025-k}, and fine-tuning \citep{meta2022human}.
However, these works often focus on idealized or low-stakes scenarios, 
and there has been very little work investigating the strategic use of language in realistic contexts. One example is \citet{ferracane-etal-2021-answer}, which examines the subjectivity involved in identifying non-sincere moves in congressional hearings. 

% Our study aims to offer a means of adapting the ME Game jury function---by extending and operationalizing it----and evaluating it over extensive, realistic, high-stakes discourse. We differ from prior work in three key ways. \textbf{First}, we extend the ME jury function by providing a parallel framing that directly incorporates Gricean maxims and proposing a commitment-based taxonomy that captures finer-grained effects of discourse moves. \textbf{Second}, to our knowledge, this is the first study to apply such theories of strategic communication to extensive, realistic strategic discourse, specifically, courtroom cross-examinations, where cooperation cannot be assumed and strategic language carries high real-world stakes. \textbf{Third}, we 
%
In our work, we apply our theoretically grounded framework
%such a framework, well-grounded in theories, 
to the study of LLMs' pragmatic understanding. Unlike prior research emphasizing downstream reasoning or task performance, our focus is on discourse understanding, a foundational layer necessary for meaningful strategic reasoning and evaluation of cooperativity.

% While our focus begins with non-cooperative discourse---where interlocutors' goals are misaligned---it is important to clarify how this connects to our framework. Non-cooperative settings are not interesting merely because they lack cooperativity, but because they make strategic language effects most salient: each move is evaluated in terms of whether it advances or undermines a speaker's goals. Our method therefore does not quantify how non-cooperative a discourse is; rather, it quantifies how strategic a move is, i.e., its payoff for the speaker. This perspective also extends to cooperative contexts, where speakers still choose strategically among possible moves, but with aligned goals. We select adversarial courtroom discourse as our testbed precisely because goal misalignment forces strategic management of commitments into the open, making it possible to study, operationalize, and evaluate.

% \section{\name: \underline{S}trategic \underline{D}iscourse \underline{A}ssessment}

\section[\name: Strategic Dialogue Assessment]%
{\name: \underline{S}trategic \underline{D}ialogue \underline{A}ssessment}

\label{SDA}
Having introduced how non-cooperativity and strategic effects of discourse moves are approached in different frameworks, we now explore two directions: (a) While still drawing on insights from the ME Game formulation of the jury function, we consider the possibility of incorporating Gricean maxims, which remain relevant for understanding goal realization in adversarial discourse and widely assumed in game-theoretic frameworks \citep{Franke2009SignalTA, goodman2013knowledge}, as an alternative way of framing the jury function.\footnote{Though we do not regard this as determinative, we do think that an advantage of using Gricean maxims in the discourse model is that they provide a relatively conservative and widely accepted account of how contextual meaning  is derived (Grice's analysis of implicatures). By contrast, models that distance themselves from Gricean maxims must rely on alternative and somewhat less widely accepted mechanisms to fulfill this role, an SDRT-like notion of coherence in the case of  ME Games. It is important to recognize that the present work does not assume Gricean cooperativity, but instead uses Gricean maxims to study levels of cooperative or pseudo-cooperative behavior in adversarial settings.} (b) We also explore potential solutions to the problem of estimating the constructs introduced by the jury function, and we attempt to capture the finer-grained effects of discourse moves observed in dialogues like that in (\ref{WMT}). By doing so, we aim to relate theory to real-world discourse and to provide a well-motivated approach to evaluating LLMs' pragmatic ability to recognize strategic effects of language.

Before turning to how we approach these two matters, we present the following dialogue (further analysed in Figure~\ref{fig:calculation}) to illustrate the components required to capture the strategic effects of language, and discuss which key assumptions and intuitions are (and are not) captured by existing approaches (e.g., ME Games). The dialogue is between the prosecutor (P) and Richard Ofshe (RO), as discussed previously. For ease of reference, we number RO's responses. The central issue here is whether the police interrogation was coercive rather than routine, a distinction that matters because coercion would undermine the reliability or admissibility of the defendant's statements. Within this exchange, the defense expert seeks to sustain the claim that the questioning exerted improper pressure, while the prosecutor aims to reframe the officers' conduct, such as asking leading questions, as standard practice and therefore 
does not amount to coercion.
%insufficient to support a coercion finding.}

% we briefly outline the assumptions we share with  ME Games in modeling discourse, as these are relevant for understanding our treatment.

\ea
\begin{itemize}
    \item[P:] [...]In what you term or what you find in there coercive [...]?
    \item[RO:] [...] The questions were very directly specifying what the answers should be. \hfill \textbf{utt1}
    \item[P:] Did you find anything in the statement [...] to indicate that the officers gave him the information about which boy was castrated?
    \item[RO:] In their statements? Perhaps there is no such record. \hfill \textbf{utt2}
    \item[P:] Ok, you also talked to Mr. Smith for three hours? 
    \item[RO:] No. I talked with him for the length of time it took to produce the transcript here. \hfill \textbf{utt3}
    \item[P:] [...] what coercive tactics do you allege that the police made in this case -- or did? .
    \item[RO:] In order to answer your question, first I need to break the interrogation down [...] so that I can cut out parts of it and focus on a particular part. \hfill \textbf{utt4}
\end{itemize}
\z

To quantify the strategic effects of these responses, we in fact evaluate what \textit{commitments} RO makes (section \ref{c-making}), and whether those commitments are meaningful, i.e., \textit{interpretable} w.r.t.\ the prosecutor's questions (section \ref{qud}). Crucially, such meaningful commitments can carry different strategic values, depending both on their \textit{content} (section \ref{content/how}) and on \textit{how they are realized} (section \ref{approx}) in discourse. We elaborate the details in the following sections \ref{c-making}-\ref{approx}.

This can be seen by comparing RO's responses across turns. In utt1, RO makes a direct and relevant commitment that characterizes the questioning as coercive; the content of this commitment supports his expert testimony and is therefore strategically \ben. In utt2, RO's response concedes (at least defeasibly) that there may be no record indicating that officers supplied the relevant information---a concession that is strategically \detr\ for RO given the prosecutor's line of attack. However, this detrimental content is conveyed with hedging (``Perhaps''), which \textit{weakens} the speaker's apparent commitment strength and can partially mitigate the loss by reducing the perceived reliability and explicitness of the concession. Utt3 has strategically \neu\ content: RO corrects an overgeneralization about the duration of the interview, which does not directly advance either side's central claim about coercion. Still, the manner of the correction may read as evasive, potentially \textit{incurring a credibility cost} despite addressing the question. Finally, utt4 does not provide an answer to the prosecutor's question at all; instead it postpones engagement by proposing a reformulation of the task, thereby failing to introduce any commitment whose content addresses the current question (\none) and incurring a strategic penalty despite offering information that may be relevant at a later stage.

These contrasts highlight three recurring dimensions of strategic evaluation. First, a response may yield an immediate benefit or penalty by advancing or undermining the speaker's goals through the content of the commitment it introduces. Second, this effect may be modulated by credibility, as it is plausible that commitments conveyed indirectly or in violation of conversational expectations often have weaker strategic impact, a dependence also reflected in the ME jury function via $\textnormal{P}_k(\textnormal{Good}_i)$. We formalize these intuitions in section \ref{sda} by defining turn-level benefits (\bat), and turn-level penalties (\pat). Third, because benefits and penalties accrue over turns, a speaker's overall strategic position depends on the cumulative balance between gains and losses. This is formalized as the normalized cumulative difference over the course of the dialogue (\nrbat).

\subsection{Discourse as commitment-making process}
\label{c-making}
A classical game-theoretic view of communication holds that language is not evaluated purely in truth-conditional terms; rather, it is viewed as a form of action. Like ME Games, we adopt the perspective that discourse is a process of making commitments \citep{walton1995commitment, farkas2010reacting, asher2017message}. As \citet{asher2017message} put it:
\begin{quote}
Crucially, some, perhaps most, of these [conversational] objectives involve commitments to contents, which are the conventional meanings and contextually derived implicatures of the utterances of the conversation. (p. 359)
\end{quote}
This implies at least two things: (1) discourse goals are achieved through the act of making commitments, and (2) more specifically, they are achieved through commitments to contents---encompassing both conventional meaning and contextually derived meaning. Fundamentally, then, the jury function measures the effects of the contents of commitments.

\subsection{Tying Goal Realization via Questions Under Discussion (QUD)}

\label{qud}
As noted above, discourse is a process of making commitments to achieve goals. Hence what we want to quantify concerns how successfully a speaker advances their own goals through such commitments. To achieve their goals, there are many possible commitments a speaker could make. However, these are not selected randomly or arbitrarily; rather, we observe particular, organized moves that are interpretable with respect to those goals. This requires some mediating principle that links individual commitments to broader discourse goals. Different accounts provide such a principle. In the ME Game jury function, goal advancement is tied to the presence of coherence relations, since coherence connects discourse structure to the goals of a text. Another option is to ground the link in the notion of Questions Under Discussion (QUD) which can be seen as a development of the notion of relevance (\citealt{roberts1996information, roberts2012information}; for overviews, see \citealt{VellemanBeaver2016QUD,beaver2017questions}). Because the current paper focuses on a type of text in which one interlocutor primarily asks questions and the other provides answers, it is natural to use the QUD framework, which views all discourse as consisting (at least implicitly) of Q/A pairs. That said, other approaches, including coherence-based accounts,  remain viable alternatives, and our choice here partly reflects both the genre of discourse and our own areas of expertise, rather than a particular piece of evidence favoring one framework over another.

In QUD-based accounts (\citealt{roberts2012information, ginzburg2012interactive}), utterances that fail to engage the current QUD are treated as \textit{non sequiturs} in the sense that they cannot straightforwardly be integrated into the discourse structure.\footnote{Failing to address the current QUD is not necessarily strategically ineffective and may serve other functions at a more global level. In this paper, however, our focus and annotations are intentionally local: the effects we measure are defined at the turn level, and we do not extend our claims beyond this level of analysis. We discuss the implications and limitations of this choice in more detail in section~\ref{limitation}.} While such utterances may still provide information in general, they do not commit to the ongoing discourse goals and therefore do not function as satisfactory strategies for advancing them (i.e., they do not maximize joint utility). QUDs thus tie discourse moves to discourse goals: only moves that address the current QUD count as contributions to goal realization, whether via  conventional meaning or contextually derived implicatures.

\subsection{Evaluating the Contents of Commitments}
\label{content/how}

We now turn to how commitments are linked to discourse goals through QUDs, specifying more precisely what makes a commitment meaningful—namely, that it is interpretable relative to the current QUD. The jury function does not simply check whether a commitment has been made; rather, it evaluates the effects of the specific content to which the speaker commits. As illustrated in example (\ref{WMT}), different commitments that address the QUD can advance the speaker's goals in different ways.

\paragraph{A Taxonomy of commitment effects}

In our system, the strategic benefits are rooted in the contents of the commitments, whether literal or implicated, that a speaker makes. Crucially, whether a meaningful commitment is strategically beneficial depends on the current QUD.\footnote{We claim that utterances are interpreted relative to the current QUD, but we do not assume that this QUD exists in isolation (e.g., the medication question is situated in a larger question of what happened in the court (rather than in the hospital)). Rather, the current QUD is part of a broader discourse structure (e.g., a hierarchy of questions that defines the context of the interaction). Our annotation scheme therefore conditions labels on the locally active QUD without denying the role of higher-level discourse organization.} For example, when the QUD is ``Are you taking any medication?" in the context of testing the witness' mental stability, a positive commitment such as ``Yes, sir" may undermine the witness's goal of maintaining credibility, while a negative commitment may support it. In addition, when the QUD is ``What is your name?"---as a confirmation question to establish that the witness is the correct person to testify---a commitment such as established by the response ``Mary" may carry no remarkable strategic gain or loss.

\begin{figure*}[t]
    \centering
    \includegraphics[width=\linewidth]{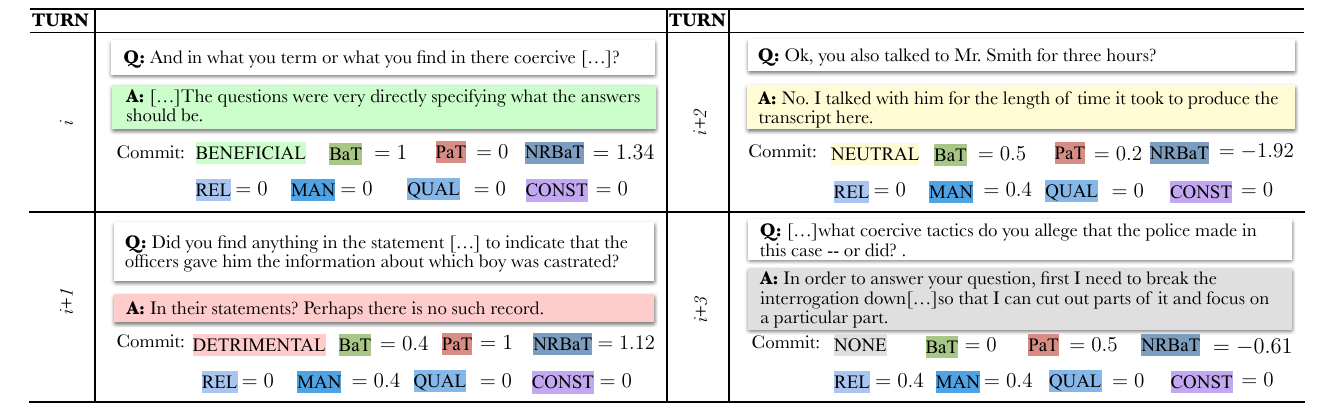}
    
    \caption{Examples of different commitment types and the corresponding values of PaT, BaT, and NRBaT}
    \label{fig:calculation}
    \vspace{-2em}
\end{figure*}

Following this logic, we propose that a commitment (\commit), relative to the current QUD, be classified as \ben, \detr, \neu\ (or impartial), or \none\ (i.e., no meaningful commitment) according their effects on the discourse goals. Figure~\ref{fig:calculation} illustrates these categories with examples from real cross-examinations. With this taxonomy, we can now capture finer-grained distinctions among the commitments in (\ref{WMT1}-\ref{WMT}). Admissions such as being rejected as an expert in another court, or acknowledging that the police did not display behaviors typical of coercion, count as \detr. By contrast, statements such as not having been refused recognition as an expert by the state, or pointing out that the police asked misleading questions, count as \ben. 

We note that this taxonomy can also be applied to the original coherence-based ME jury function in order to capture finer-grained effects, as we have done here. Specifically, one can assume that the absence of a coherence relation functions like \none\ in our categorization: it ties discourse moves to discourse goals, while still allowing for distinctions in the effects of coherent moves. We view this as a promising direction for those interested in a coherence-based approach, though the literature also notes practical obstacles in annotating coherence relations compared to Gricean maxims \citep{hoek-scholman-2017-evaluating, alkorta-etal-2019-towards, sanders2021unifying, panzeri2021children}.

To summarize, we have presented one way of modeling how a dialogue is formed and interpreted, drawing on Grice's analysis of implicatures together with QUD-based discourse structure. More importantly, we have introduced a taxonomy that links each linguistic move to its contribution to the speaker's overarching goals.

\subsection{Coping with Partially Observable Terms}
\label{approx}

Having settled how the foundational commitment effects are determined, we now turn to other terms that modify them, namely, $\textnormal{P}_k(\textnormal{Good}_i)$, $\textnormal{win}_{i}{(k)}$, and {\sc cons} according to Eq.\ref{tau}. These three terms describe, respectively: how reliable a commitment is, whether the commitment eliminates winning potential, and whether the commitment contradicts previous commitments. They model the extent to which the commitment effects should be preserved. For example, if a beneficial commitment is unreliable, the benefits the speaker gains should be proportional to the speaker's credibility. If a detrimental move eliminates all possible winning paths, then no further benefits will be awarded. Similarly, if a commitment contradicts a prior beneficial one, the benefits gained from the latter should be retracted. Of the three, consistency is straightforward and requires no further specification, whereas the concept of credibility and winning potential are not directly observable from discourse alone. In our treatment, we approximate a speaker's credibility through violations of Gricean maxims, while leaving the modeling of winning potential aside. The win function  distinguishes moves that automatically lead to loss from those that do not, and we are not aware of any way of identifying this property, at least in our dataset. There would be further challenges if these partially observable terms were to be incorporated in our model directly, as we discuss in section~\ref{limitation}.

\begin{wrapfigure}{r}{0.5\textwidth}
    \begin{minipage}{\linewidth}
    \centering
    \includegraphics[width=0.9\linewidth]{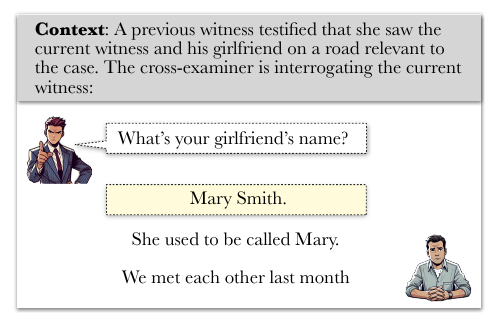}
    \caption{An example illustrating the role of violations of maxims as diminishing the reliability (thus also the effect size) of the commitment.}
    \label{fig:motivating_ex}
    \end{minipage}
\end{wrapfigure}
\paragraph{Approximating $\textnormal{P}_k(\textnormal{Good}_i)$} Violations of maxims are assumed to link to the speaker's credibility, typically causing a decrease in credibility \citep{shuy1998language, ginzburg2012interactive}. Consider an example from a cross-examination in Figure \ref{fig:motivating_ex}.The prosecutor presses the defendant to name his girlfriend, perhaps to put the information on the record, or to distract him. We present three alternative answers. All three options are classified as neutral commitments, yet the final effects they produce differ. The two alternatives that violate maxims---such as those of manner and relevance---can be perceived as less credible, thereby appearing diminished in effect compared to commitments that involve no violations.\footnote{That said, commitments that violate maxims are perceived as weaker versions of their type: less beneficial, or less detrimental than comparable commitments without violations.} 

We note that maxim violations do not constitute the only factors influencing the interlocutor's credibility distribution, but they provide observable cues that feed into credibility assessments. In addition, they describe how a meaningful commitment is realized, distinguishing explicit commitment-making from implicit commitment-making via implicatures.
In this respect, maxim violations play a role analogous to $\textnormal{P}_k(\textnormal{Good}_i)$ in the original function: they modulate the base effects (from the contents of the commitment) by affecting the perceived reliability of the commitment. Accordingly, we decompose the function from speakers to their credibility distributions into violations of relevance, manner, and quality.\footnote{One might ask why we still retain \textsc{cons} later, given its apparent overlap with the maxim of quality. We keep \textsc{cons} because it targets overt self-contradictions, whereas credibility assessments (reflected in $\textnormal{P}_k(\textnormal{Good}_i)$) capture subtler forms of epistemic doubt. In this proposal, we treat violations of the maxim of quality as gradable. This is motivated by practical concerns: in real-world discourse, truth is often not directly accessible, and what we can observe are degrees of deception or implausibility rather than categorical falsehoods. For the same reason, we conflate the maxims of quality and quantity, treating both as contributing to the hearer's plausibility assessment.} When no maxim is violated, the speaker appears to be cooperative (\citealt{levinson2000presumptive, horn2004}), which can serve as a strategy to minimize the loss due to credibility.

\subsection{Underlying Elements of \bfname{}} 

\label{sda}

Up to this point, we have provided a way to incorporate Gricean maxims into the jury function, in parallel to the ME Game formulation; to capture the finer-grained effects of discourse moves; and to develop a more estimable account of the constructs introduced by the jury function. We call our approach Strategic Dialogue Assessment (\name), and now elaborate on its scoring mechanism, our variant of the ME Games jury function. We locate base strategic value primarily in the impact of commitment contents on discourse goals. 
%We name our approach Strategic Dialogue Assessment (\name) and we now elaborate on the \name\ scoring mechanism, which supplants the ME games jury function. 
%We directly take effects of contents of commitments on discourse goals as the primary source of base strategic value. 
Based on their effect on the speaker's goals, we categorize commitments (\commit) into four types with different assigned values: \ben, \detr, \neu\ (or impartial), and \none.

In addition, we replace $\textnormal{P}_k(\textnormal{Good}_i)$ with penalizing/rescuing terms based on violations of the Gricean maxims of relevance, manner, and quality. Specifically, violations of relevance (\rel) and manner (\man) are treated as \textit{multiplicative} terms (see Eq.\ref{eq1}). When a witness makes a beneficial/neutral commitment in an irrelevant or unclear statement, the strategic gain is diminished compared to a response that is relevant and clear. Detrimental commitments conveyed through implicature are penalized less severely, since they avoid explicitly harming the speaker's interests. In such cases, the indirectness provides some strategic compensation.\footnote{Note that such compensation does not apply to \none, as no meaningful (i.e., interpretable) commitment is made in the first place (e.g., when the response has strongly violated relevance and/or manner).} Truthfulness (quality; \qual) is also modeled as a \textit{multiplicative} factor (see Eq.\ref{eq1}), capturing whether the speaker is still perceived as trustworthy and sincere from prior turns. Together, these refinements extend and decompose $(\textnormal{COH}_i(k) + \textnormal{RES}_i(k)) \times \textnormal{P}_k(\textnormal{Good}_i)$ into interpretable discourse properties: commitments and violations of the Gricean maxims.

Finally, we maintain a constraint on consistency (\const), which reflects a key pragmatic pressure: speakers generally avoid inconsistent commitments, even if doing so requires conceding harmful facts or adopting strategically ambiguous responses. This pressure is further reinforced in legal settings by the potential availability of impeachment, which targets inconsistencies in a witness's statements. Unlike \citet{asher2017message}, however, we do not assume that inconsistency leads to a total collapse in strategic standing or nullifies future benefits. Instead, we treat inconsistency as a strong, but not absolute, penalty that significantly reduces the value of current strategic gains. This models the intuition that inconsistency undermines the reliability of prior contributions. Taken together---\commit, maxim violations, and consistency---these are the factors that we take to contribute to the overall effect of a commitment.

As should be clear, while our approach is heavily inspired by the ME Game jury function, we do not directly reproduce it. Instead, we incorporate elements such as QUDs and Gricean maxims, which provide a better adaptation for the type of discourse we analyze and offer a relatively estimable way to model speaker credibility. We leave it open whether future research could develop an even tighter connection to the original ME Game formulation.

\paragraph{PaT, BaT and NRBaT}
We define a value assignment function for commitments $f_c$, which maps the commitment $C_i$ at turn $i$ to a corresponding score. For simplicity, we assign a value of $1$ to beneficial commitments and $-1$ to detrimental ones. Neutral commitments are treated as carrying a weak positive benefit, while the absence of meaningful commitment is treated as a penalty, reflecting the pragmatic pressure that commitments should address the current QUD in order to be interpreted as contributing to discourse goals. Since this requirement concerns whether a commitment is meaningful---that is, capable of contributing to discourse goals---rather than being a direct contributor to discourse goals itself, we treat its effects as weaker than those of beneficial or detrimental commitments: 
% \begin{small}
\begin{align}
    f_c(C_i) = \left\{
    \begin{array}{ll}
    \phantom{-}1 & \textnormal{if $C_i = \ben$} \\
    \phantom{-}0.5  & \textnormal{if $C_i = \neu$} \\
    -0.5  & \textnormal{if $C_i = \none$}\\
    -1  & \textnormal{if $C_i = \detr$}
\end{array}
\right.
\end{align}
% \end{small}
Thus, the Benefit at Turn $i$ (\bati)\footnote{A weighted sum is in theory possible with this equation; however, since we are interested in measuring correlation rather than absolute values, we leave this for future work. There are many ways to combine commitment types and violations of dialog conventions into a scalar score, and learning a function directly from outcomes can be an alternative. However, instead of relying on theoretically identified discourse properties, a fully learned function could obscure these distinctions by collapsing multiple pragmatic factors into uninterpretable weights. Exploring hybrid approaches that combine theoretical structure with learned parameters is an important direction for future work.} and the Penalty at Turn $i$ (\pati) can be computed as:\footnote{While our computation treats violations as binary labels (presence or absence), our annotation process used a finer-grained 4-point scale. In this scale, scores of 1–2 were treated as absence (no violation or borderline cases), while scores of 3–4 were treated as presence (clear and strong violations). This finer scale was used to reduce potential ambiguity introduced by forcing annotators to make binary choices during labeling, before collapsing the labels into a binary form for computation.}

\begin{align}
\label{eq1}
    \textnormal{BaT}_i    &= \left\{
    \begin{array}{ll} f_c(C_i), \textnormal{if}\,C_{i} \in \{\ben,\neu\}\\
     f_c(C_i) \times (\textnormal{Rel}_i + \textnormal{Man}_i + \textnormal{Qual}_i), \\ \hspace{3em}\textnormal{if}\,C_{i} = \detr\\
    0,\hspace{2em}\textnormal{otherwise}
    \end{array}
    \right.\\
    \textnormal{PaT}_i    &= \left\{
    \begin{array}{ll}
    |f_c(C_i)|  + \textnormal{Const}_i \times \sum_{j=1}^{i} \textnormal{BaT}_j, \\ \hspace{3em} \textnormal{if}\,C_{i} \in \{\detr,\none\} \\
    |f_c(C_i)| \times (\textnormal{Rel}_i + \textnormal{Man}_i + \textnormal{Qual}_i) \\
    \hspace{1 em}+\, \textnormal{Const}_i\times\sum_{j=1}^{i} \textnormal{BaT}_j,\, \textnormal{otherwise}\\
    \end{array}
    \right.
\end{align}

In addition, we define the Normalized Relative Benefit at Turn $i$ (\nrbati) to capture the cumulative, normalized relative benefits across discourse: 
\begin{small}
\begin{align}
    \textnormal{NRBaT}_i  &= Z\left(\sum_{j=1}^{i} \textnormal{BaT}_j\right) - Z\left(\sum_{j=1}^{i} \textnormal{PaT}_j\right)
\end{align}
\end{small}

\noindent This formulation computes the cumulative sums of BaT and PaT and applies a z-score normalization to these sums to ensure comparability between gains and penalties. The difference between the normalized BaT and normalized PaT at turn $i$ provides an estimate of the overall strategic value accumulated over the discourse up to that turn. To avoid confusion, we note that the cumulative score aggregates \textit{local} strategic effects under the assumption that immediate commitments and credibility shifts matter incrementally. However, it is  not intended to capture long-horizon strategic planning or delayed traps, that is, moves for which the strategic benefit only materializes later. Capturing delayed strategic payoffs would require explicit modeling of goal hierarchies and future contingencies, which lies beyond the scope of the present work. Figure~\ref{fig:calculation} presents the corresponding values of our metrics for an adapted snippet from a real cross-examination, with a detailed example of the calculation provided in Appendix \ref{cal_steps}. We accept, of course, that the particular numeric values we used in the above definitions of scoring functions are theoretically somewhat arbitrary. We leave open whether in future work (a) a ranking method could be used that obviates the need to specify particular numeric values, or (b) a learning procedure or theoretical argument could be used to identify appropriate numeric values.

% \Needspace{5\baselineskip}

\section{Human Annotations on \dataset}
\label{annotation_guideline}

\begin{wrapfigure}{r}{0.5\textwidth}
    \begin{minipage}{\linewidth}
    \centering
    \includegraphics[width=0.8\linewidth]{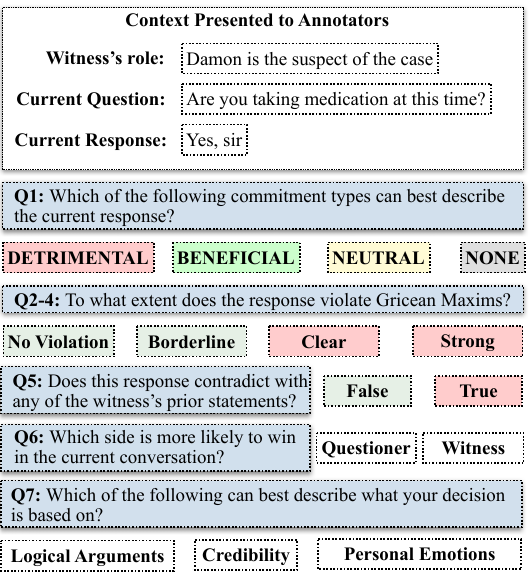}
    \caption{Annotation schema (guideline in Appx.~\ref{protocol}).}
    \label{fig:annotation_schema}
    \end{minipage}
    \vspace{-3em}
\end{wrapfigure}

To empirically evaluate our framework and assess LLMs' pragmatic abilities, we conduct human annotations on our dataset \dataset. By applying \name\ to \dataset, we reveal that the identified discourse properties and the derived metrics in fact capture how non-cooperative discourse differs from its cooperative counterpart and we validate the strategic effects measured by the function are meaningful via an outcome prediction task.

\paragraph{Human Annotations}

We conduct human annotation on a subset of our dataset, covering approximately 200 turns per side per trial, for a total of around 800 turns. We recruit three annotators with relevant linguistic expertise (a journalist and two linguistics students) via Upwork, offering a pay rate of \$20 per hour. The annotators first complete a pilot set to familiarize themselves with our annotation framework, followed by a shared aggregated set consisting of a complete cross-examination (of around 150 turns). Given that reliable inter-annotator agreement is seen within this shared set, we proceed with them to annotate separate cases across different trials, balancing resource limitations with the goal of achieving broad dataset coverage.

\begin{wraptable}{r}{0.5\textwidth}
  \vspace{-0.6em}
  \caption{Inter-annotator agreement statistics.}
  % \vspace{-0.6em}
  \centering
  \scriptsize
  \begin{tabular}{@{}l l l@{}}
    \toprule
    \multicolumn{3}{@{}l}{\textit{Metric-level agreement}} \\
    \midrule
    \bat (Spearman's $\rho$) & \textbf{0.65} & \\
    \pat (Spearman's $\rho$) & \textbf{0.66} & \\
    \nrbat (Spearman's $\rho$) & \textbf{0.83} & \\
    \commit(Fleiss' $\kappa$) & \textbf{0.59} & \\
    \rel (Randolph's $\kappa$)\tnote{a} & \textbf{0.72} & \\
    \man(Randolph's $\kappa$)\tnote{a} & \textbf{0.52} & \\
    \qual(Randolph's $\kappa$)\tnote{a} & \textbf{0.86} & \\
    \const (avg.\ TPR)\tnote{b} & \textbf{25\%} & \\
    \midrule
    \multicolumn{3}{@{}l}{\textit{Outcome-level agreement}} \\
    \midrule
    Outcome decision (Fleiss's $\kappa$)\tnote{c} & \textbf{0.29} & \\
    Jaccard similarity (avg.\ reasons) & \textbf{0.46} & \\
    Complete agreement on reasons & \textbf{29\%} & \\
    \bottomrule
  \end{tabular}
  \vspace{-1em}
\end{wraptable}
Figure~\ref{fig:annotation_schema} illustrates the annotation task. Annotators read through a cross-examination dialogue in temporal order. For each turn, they are presented with background information (e.g., the witness's role) and the current Q/A pair. They are then asked to evaluate the response along three dimensions: type of commitment, violation of each of the Gricean maxims, and consistency (see our protocols in Appendix ~\ref{protocol}). Additionally, we elicit annotations on more basic and widely used indicators of strategic behavior, such as outcome judgments \citep{duan2024gtbench} and the underlying reasons for those outcomes following \citet{lukin-etal-2017-argument, sep-aristotle-rhetoric, xu-etal-2024-earth}, which serve as a testbed for evaluating the applicability of our framework.

% Inter-annotator Spearman's $\rho$ values for BaT, PaT, and NRBaT are \textbf{0.65}, \textbf{0.66}, and \textbf{0.83}, respectively. We also report inter-annotator agreement on the subcategories: Cohen's $\kappa$ for commitment value is \textbf{0.59}; Randolph's $\kappa$\footnote{Randolph's $\kappa$ \citep{randolph} is used for violation categories because their label distribution is usually highly skewed (see also Figure~\ref{fig:maxim}).} for violations of relevance, manner, and quality are \textbf{0.72}, \textbf{0.52}, and \textbf{0.86}, respectively; and average true positive rate for consistency is \textbf{25\%}.\footnote{Inconsistencies are rare and typically signal obvious lying, so we report only the average true positive rate.} In contrast, the average inter-annotator agreement on outcome decisions,\footnote{We note that our multiple regression analysis predicts outcomes on a per-annotator basis, so the observed disagreement on outcome among annotators does not contradict the good predictive power of our metrics.} measured by Cohen's $\kappa$, is merely at \textbf{0.29}, indicating low agreement. Even in cases where all three annotators reached the same decision, their rationales showed limited overlap: the average Jaccard similarity among selected reasons was 0.46, and only 29\% of such cases exhibited complete agreement on reasoning. Second, 

We find that annotators exhibit high agreement\footnote{We note that there is no universally agreed threshold for “good” inter-annotator agreement in pragmatics annotation. However, prior work in discourse and pragmatics typically treats $\kappa \approx$ 0.5-0.7 as reasonable, given the inherently interpretive nature of the task \citep{artstein-poesio-2008-survey, hoek-scholman-2017-evaluating}.} on our metrics, while other metrics such as outcomes are much more subjective. We assess inter-annotator agreement using several metrics. For BaT, PaT, and NRBaT, we report Spearman's $\rho$ values. For the subcategories, we use Fleiss' $\kappa$ to measure agreement on commitment types, and Randolph's $\kappa$\footnote{Randolph's $\kappa$ \citep{randolph} is used for maxim violations because their label distribution is usually highly skewed (see also Figure~\ref{fig:maxim}).} to evaluate violations of relevance, manner, and quality. For consistency, we report the average true positive rate.\footnote{Inconsistencies are rare and typically signal obvious lying, so we report only the average true positive rate.} Finally, for outcome decisions we use Fleiss' $\kappa$,\footnote{We note that our multiple regression analysis predicts outcomes on a per-annotator basis, so the observed disagreement on outcome among annotators does not contradict the good predictive power of our metrics.} and we additionally examine the overlap in annotators' rationales using Jaccard similarity and the proportion of cases with complete agreement. We highlight here that annotator agreement is consistently higher for our metrics and discourse properties than for conversational outcomes.

\section{Non-cooperative and Cooperative Discourses Are Asymmetric}
%make it into a section conclusion; boost up to the top, giving you some motivated way of annotating; a corpus of the annotations; to measure strategic effects of both cooperative and non-cooperative discourse; we report on the followings; not test theories; in revealing for showing what the gaps between theories and real discourse; it's useful to demonstrate the value of our methods in llms 
%(CPI)crooked path to innocence corpus.

Criminal trials offer a convenient setting in which to distinguish cooperative from non-cooperative discourse. In direct examination, a lawyer questions their own witness; here the lawyer's and witness's goals can be taken to be aligned, so the exchange is cooperative. In cross-examination, by contrast, a lawyer questions the opposing side's witness; here the goals are non-aligned, so the exchange is non-cooperative. Comparing these two types of examination, we find an asymmetry in the strategic effects that discourse moves convey.

\begin{wrapfigure}[23]{r}{0.5\textwidth}

    \begin{minipage}{\linewidth}
    \centering
    \includegraphics[width=\linewidth]{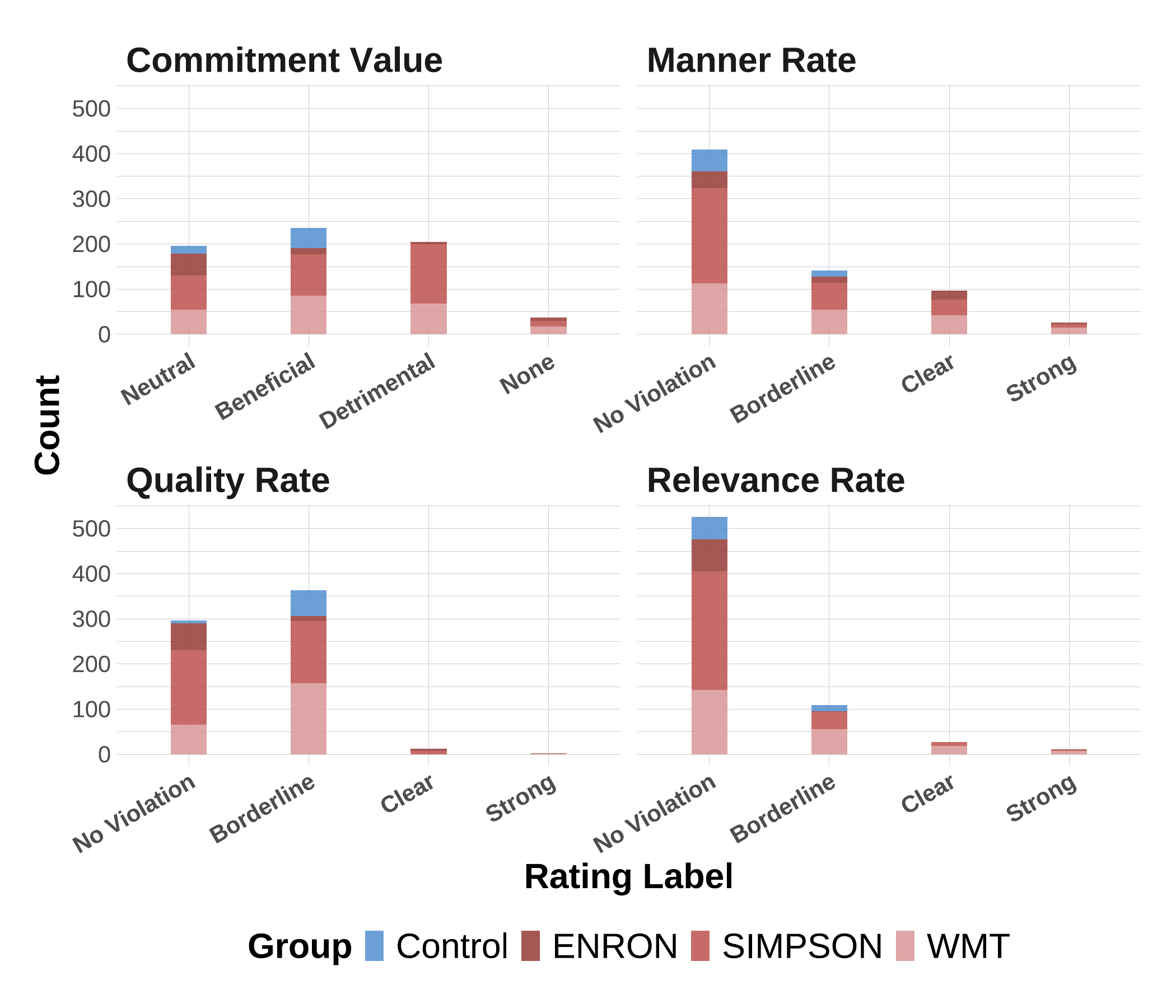}
    % \vspace{-2.5\baselineskip}% DIBHACK
    \caption{Distribution of Gricean maxim ratings and commitment types across trials}
    \label{fig:maxim} 
    \end{minipage}
    \vspace{-3em}
\end{wrapfigure}

Figure~\ref{fig:maxim} shows the frequencies of different commitment types and maxim violations in both settings. Control discourse (the blue bars) rarely involves detrimental commitments or violations of maxims, whereas non-cooperative discourse (the reddish bars) exhibits a higher frequency of both phenomena. 
% \begin{wrapfigure}{r}{0.5\textwidth}
%     \begin{minipage}{\linewidth}
%     \centering
%     \includegraphics[width=0.9\linewidth]{pics/motivating_example.png}
%     \caption{An example illustrating the role of violations of maxims as diminishing the reliability (thus also the effect size) of the commitment.}
%     \label{fig:motivating_ex}
%     \end{minipage}
% \end{wrapfigure} 
% \Needspace{16\baselineskip}% DIBHACK

Our next observation is that even within non-cooperative discourse, the occurrence of maxim violations is remarkably lower than that of maxim maintenance, suggesting that violations alone may not sufficiently capture the strategies used in non-cooperative discourse. By incorporating the commitment taxonomy, \name{} more accurately represents non-cooperative discourse and allows for more nuanced interpretations of violations, for instance, as loss minimization or benefit retrieval strategies.

\begin{wrapfigure}{r}{0.45\textwidth}
    \begin{minipage}{\linewidth}
    \centering
    \includegraphics[width=\linewidth]{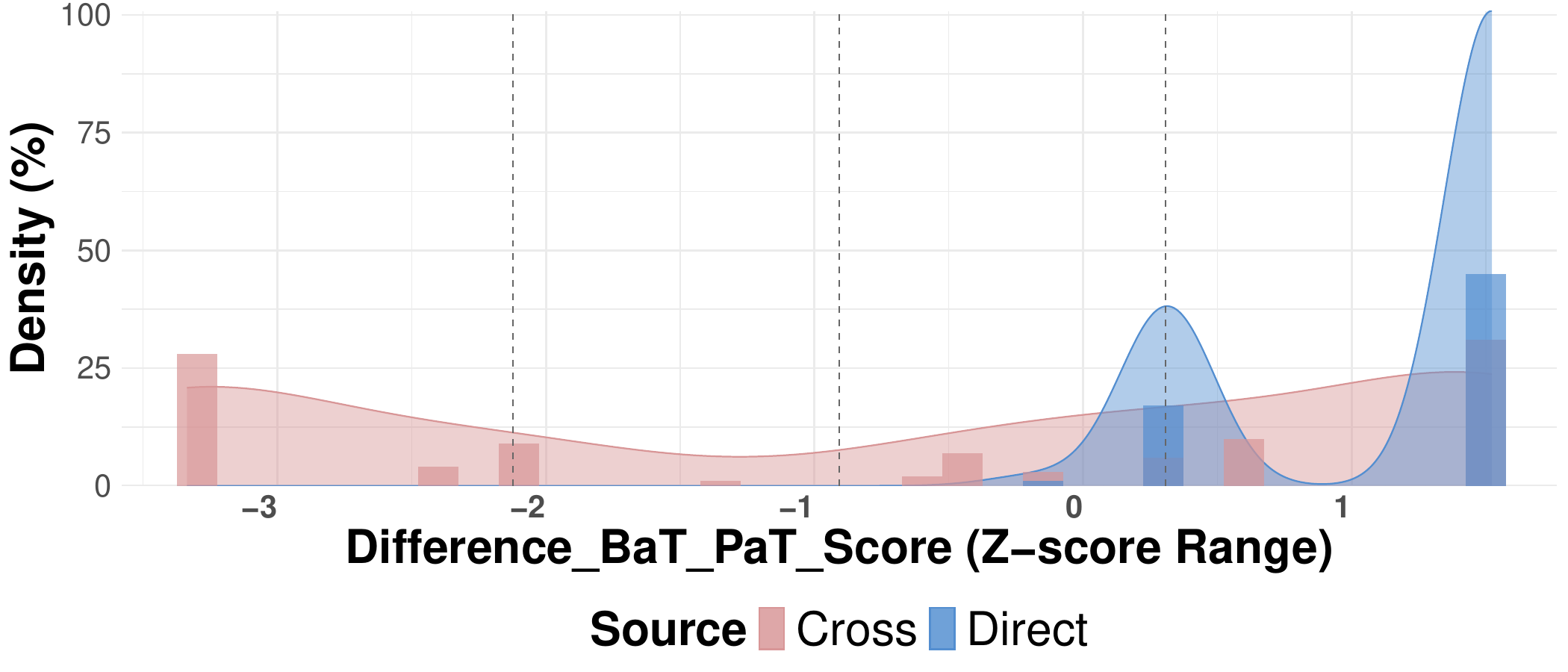}
    \caption{Distribution of net move benefit (PaT-BaT, z-scored) across discourse types. Control (cooperative) discourse concentrates around high benefit values, while cross-examination (non-cooperative) discourse shows a wider spread, reflecting greater variability and tension between gains and losses.}
    \label{fig:control}
    \end{minipage}
    \vspace{-2em}
\end{wrapfigure}
Having seen that when comparing cooperative and noncooperative dialogue there are differences both in frequencies of different commitment types, and in frequencies of maxim violations, we now show that \name{} metrics (i.e., BaT, and PaT), like the underlying local discourse properties, also distinguish cooperative from non-cooperative discourse. As illustrated in Figure~\ref{fig:control}, we use the z-scored difference between PaT and BaT at each turn to capture net move benefit (with NRBaT representing the cumulative counterpart). The results show that control discourse primarily exhibits benefits (i.e., more likely to win), while the non-cooperative counterpart displays a wider distribution, highlighting the inherent tension created by the need to appear cooperative during an inherently adversarial interaction. 

% We acknowledge that due to resources limitations, we were not able to collect extensive annotations on cooperative discourse, and how much the results from the comparison we made here can be extended to give a more conclusive conclusion on differece between discourse types is 

Before moving on, we must note a limitation of the above results, namely that our comparison between cross-examination and direct examination (i.e., control) conversations is based on a single witness for whom we collected both transcripts. For all other cases, we annotated only the cross-examinations, which explains the disparity in the number of examples across the two discourse types. Since our primary interest lies in non-cooperative discourse rather than in providing a systematic overview of discourse typology, this single instance of direct examination serves as a sufficient contrastive baseline. Moreover, given the time-intensive nature of the annotation task,we chose to concentrate our efforts on cross-examinations. Accordingly, the results of this comparison should be viewed as illustrative rather than conclusive: they highlight potential contrasts between discourse types, but more extensive data gathered in future work would be needed to establish firm generalizations. However, even absent data from a large sample, we feel the differences between cooperative and non-cooperative discourse we have discussed in this section provide an important proof-of-concept for the general approach we have adopted, demonstrating how such an approach can be used to distinguish different types of discourse.

\section{Intrinsic Evaluation}
\label{intrinsic eval}
\name\ offers a way to integrate Gricean maxims and commitment-based taxonomy into the original ME Game jury function, with the goal of modeling the strategic effects conveyed by a discourse move. We have provided justification for our pragmatic choices, arguing that our treatment offers a practical and valid extension of the original ME Game jury function. In this section, we complement that with empirical evidence from human annotations, demonstrating that our framework is well-suited to capturing the dynamics of real-world discourse.

We present our findings from three perspectives. (1) Similar to the original jury function, which could point toward the outcome of a game, our modified metrics are also predictive of conversational outcomes. This suggests that our modifications preserve key insights from the original function regarding how utterances contribute to outcomes (i.e., the realization of discourse goals). (2) Conditioning on different reasons for outcome judgments, we show that our metrics better capture the objective components of decision-making. (3) Compared to existing methods for quantifying strategic language understanding of LLMs---such as the NRA (Normalized Relative Advantage) proposed by \citet{duan2024gtbench}---our metrics show greater robustness to subjectivity among annotators. 

% 1) We observe a clear asymmetry between cooperative and non-cooperative discourse in terms of the frequency of maxim violations, the distribution of commitment types, and, by extension, the behavior of our proposed metrics. 

% We present three key findings: (1) Non-cooperative discourse differs from cooperative discourse, and this distinction is quantitatively reflected in the discourse properties we identify: violations of maxims and commitments---and consequently in our proposed metrics. (2) Our metrics are effective in predicting conversational outcomes, indicating that they capture meaningful strategic patterns. (3) Compared to outcome-based metrics such as NRA \citep{duan2024gtbench}, our metrics show greater robustness to subjectivity among annotators, and better reflect the objective components of the decision-making. 
%\subsection{A Comparison to Other Metrics}

\subsection{PaT and BaT are Effective Predictors of Individual Annotations of Outcome} 
\begin{wraptable}{r}{0.58\textwidth}
  \vspace{-0.6em}
  \caption{LR summary (BaT$+$PaT $\to$ outcome)}
  %\vspace{-0.6em}
  \centering
  \scriptsize
  \begin{tabular}{@{}l
                  S[table-format=+1.2]
                  S[table-format=1.2]
                  S[table-format=1.2]
                  l
                  l@{}}
    \toprule
    Predictor & {$\beta$} & {SE} & {OR} & {95\% CI } & {$p$} \\
    \midrule
    BaT & 1.47 & 0.34 & 4.33 & [0.80, 2.15] & $<$.001 \\
    PaT & -1.77 & 0.35 & 0.17 & [-2.46, -1.09] & $<$.001 \\
    \midrule
    \multicolumn{6}{@{}l}{\textit{Model fit}}\\
    \multicolumn{6}{@{}l}{AIC = 332.9 (cf. intercept-only = 419)}\\
    \multicolumn{6}{@{}l}{Tjur's $R^2$ = 0.28; Accuracy = 74.6\%; AUC = 0.80 [0.75, 0.85], $p < .0001$}\\
    \midrule
    \multicolumn{6}{@{}l}{\textit{Bootstrap (1{,}000 resamples)}}\\
    \multicolumn{6}{@{}l}{BaT: 95\% CI = [0.69, 2.15]; PaT: 95\% CI = [-2.58, -1.05]}\\
    \bottomrule
  \end{tabular}
  \vspace{-1em}
\end{wraptable}
In theory, the jury function evaluates the strategic gains and losses of each turn, which collectively contribute to, though not completely determine, the overall outcome of the discourse. Therefore, our proposed metrics, designed as an operationalizable approximation of the ME Game jury function, should also exhibit predictive power for conversational outcomes. We note that outcome judgments are inherently subjective and annotators often disagree, but they are not arbitrary: each annotator's judgments are systematically shaped by the discourse properties our metrics capture. Thus, the point of prediction is not to recover a single ground truth outcome, but to validate that our metrics track the factors that drive human reasoning about outcomes. To empirically validate this point, we conduct a basic regression analysis on our annotation data. We emphasize that this evaluation is \textbf{non-trivial}, as no prior work has empirically demonstrated that the theoretically identified formulation in the ME Game jury function necessarily reflects the achievement of conversational goals in extensive real discourse data. We conduct a multiple logistic regression analysis to predict \emph{each annotator}'s outcome judgment at each turn, using the corresponding BaT and PaT scores as independent variables.

The overall model was statistically significant and demonstrated good fit to the data. BaT was a significant positive predictor, and PaT was a significant negative predictor. The model correctly classified 74.6\% of cases, with an AUC of 0.80, indicating BaT and PaT have good discriminative ability in outcome prediction. We validated the robustness of our logistic regression results using a non-parametric bootstrap procedure with 1000 resamples.

% {\color{gray}(AIC = 332.9;\footnote{cf. the intercept-only model with an AIC of 419.} Tjur's $R^2$ = 0.28)}, {\color{gray}($\beta$ = 1.47, SE = 0.34, OR = 4.33, 95\% CI [0.80, 2.15], $p < .001$)}, {\color{gray}($\beta$ = -1.77, SE = 0.35, OR = 0.17, 95\% CI [-2.46, -1.09], $p < .001$)}, {\color{gray}(95\% CI [0.75, 0.85], $p < .0001$)}, {\color{gray}(BaT: 95\% CI = [0.69, 2.15]; PaT: 95\% CI = [-2.58, -1.05])}

Given that one of our purposes later is to evaluate LLMs' \textbf{intrinsic} ability to understand strategic language, we also experimented with LLM-as-judge (zero-shot) and found that even the best-performing model (AUC = 0.68) is outperformed by predictions based on our metrics (see detailed scores in Appx.\ Table \ref{tab:model_auc}). This indicates that the intrinsic ability of LLMs to understand the utterances and their effects on outcome is less effective than when assessed through \name{} metrics. While we expect there are ways to improve LLMs' performance on this task (e.g., through fine-tuning using our metrics), outcome prediction is not our primary focus. Rather than attempting to optimize performance on this task, we use it solely as a means of validating our proposal and its role as an evaluation metric for LLM benchmarking.

\subsection{\name{} Captures Objective Aspects of Outcome Judgments}
\begin{wrapfigure}{r}{0.55\textwidth}
    \begin{minipage}{\linewidth}
    \centering
    \includegraphics[width=0.8\linewidth]{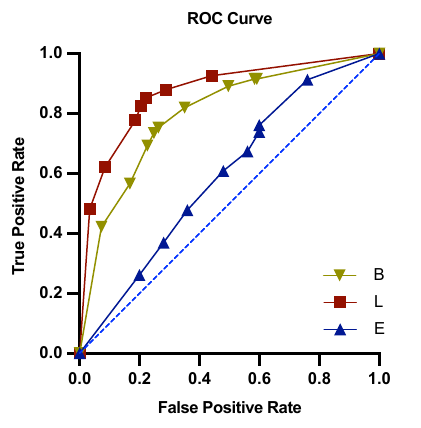}
    \caption{\small Model performances conditioned on different outcome reasons. B: baseline (i.e., without any conditions), L: conditioned on logical arguments, and E: conditioned on personal emotions.}
    \label{fig:condition}
    \end{minipage}
    % \vspace{-6em}
\end{wrapfigure}

Our metrics reflect decision-making driven by what can reasonably be thought of as objective reasons, such as logical argumentation and certain aspects of credibility building, but not personal emotions.\footnote{Note that the distinction between categories \textit{logical argumentation}, \textit{credibility building} and \textit{emotional}, as well as corresponding explanations for each category, are drawn from \citet{xu-etal-2024-earth}. To exemplify how we apply these distinctions,} a witness's admission of taking medication may objectively raise concerns about their mental state---an inference grounded in logical reasoning. In contrast, discrediting a witness solely because they do not attend church, as might occur with a biased juror, reflects a subjective and emotionally driven judgment. We fit two separate multiple logistic regression models: one conditioned on outcome decisions attributed to logical arguments and another on those influenced by personal emotions. The results in Figure \ref{fig:condition} show that the discriminative power of our metric increases significantly when 
the reasons stated for the annotated outcome are logical arguments ($p < .0001$), but drops substantially when they are personal emotions ($p < .05$). These findings further corroborate that outcome judgments are inherently subjective, whereas \name{} metrics capture the more objective aspects of outcome evaluation.

We note two important points based on these results. (1) Our operationalization of the ME Game jury function captures the contributions a speaker makes toward realizing discourse goals, but it does not determine the final outcome.\footnote{For instance, even if a witness makes many detrimental moves (as in \ref{WMT1}–\ref{WMT}), the jury may interpret this as evidence that the witness is less reliable or even ``guilty" of certain activities. Yet the final outcome depends on many additional factors, such as the presence of direct evidence of police coercion, or broader considerations of credibility and bias.} Accordingly, it is expected that the effects we capture represent only a partial account of how outcomes are decided. (2) While our metrics are designed to be relatively objective, we acknowledge that some degree of subjectivity remains, as indicated by levels of inter-annotator agreement. For example, annotators may differ in their interpretation of utterances and their effects on the speaker's goals, leading to different assignments of commitment types to the same utterance. Such variability is unavoidable, but our results suggest that our approach has minimized these subjective elements sufficiently to serve as a reasonably objective and scalable evaluation method.

\subsection{\name{} Metrics Are Consistent}
\label{objective}
In the literature, there are other metrics that assess LLMs' strategic understanding, but these have been developed for competitive games with formal payoff structures rather than for real-world discourse. One such metric is the Normalized Relative Advantage (NRA) introduced by \citet{duan2024gtbench}. NRA was designed to evaluate relative performance in settings such as poker and auctions, providing a normalized measure of wins and losses between two agents. To the best of our knowledge, it has not been previously applied to free discourse outside of a game setting. Formally, NRA is defined as the difference between the cumulative number of wins for the witness and the questioner up to turn $i$, normalized by the total number of scoring events ($\textnormal{NRA}_i = \frac{\sum_{j=1}^i \textnormal{win}_w^j - \sum_{j=1}^i \textnormal{win}_q^j}{\sum_{j=1}^i \textnormal{win}_w^j + \sum_{j=1}^i \textnormal{win}_q^j}$).\footnote{We follow \citet{duan2024gtbench}, assigning a value of 1 for a win and 0 otherwise.}

%In the literature, there are other metrics that assess LLMs' strategic understanding, but they focus purely on outcomes rather than on the finer-grained pragmatic effects on outcomes as we did. One is the Normalized Relative Advantage (NRA) from \citet{duan2024gtbench}. NRA is defined as the difference between the cumulative number of wins for the witness and the questioner up to turn $i$, normalized by the total number of scoring events ($\textnormal{NRA}_i = \frac{\sum_{j=1}^i \textnormal{win}_w^j - \sum_{j=1}^i \textnormal{win}_q^j}{\sum_{j=1}^i \textnormal{win}_w^j + \sum_{j=1}^i \textnormal{win}_q^j}$).\footnote{We follow their work: 1 for a win and 0 otherwise.} 

%\paragraph{\name{} metrics are more consistent} 
While both NRBaT and NRA aim to capture cumulative benefit over time, we observe\begin{wrapfigure}{r}{0.5\textwidth}
    \begin{minipage}{\linewidth}
    \centering
    \includegraphics[width=0.9\linewidth]{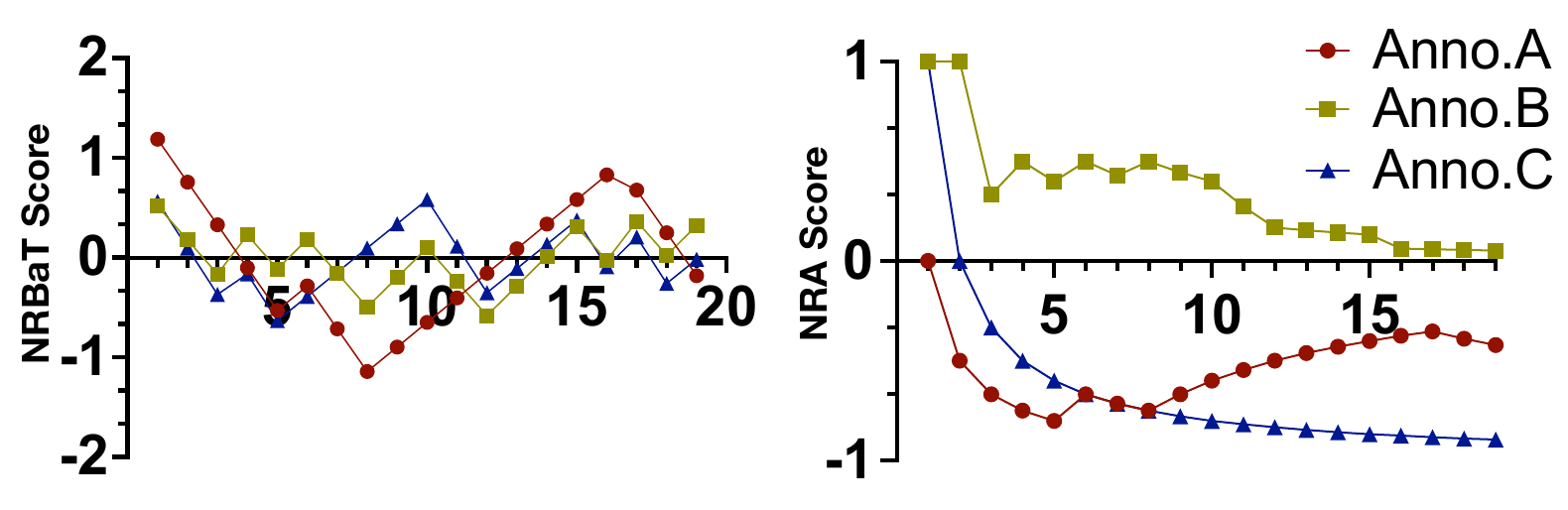}
    \caption{NRBaT and NRA of three annotators across turns during a line of questioning about medication}
    \label{fig:NRBaT-pilot}
    \end{minipage}
    \vspace{-1em}
\end{wrapfigure} that NRA is more variable. Figure~\ref{fig:NRBaT-pilot} illustrates this in a sequence of annotations concerning a line of questioning about medication. Compared to NRBaT, NRA fluctuates more sharply, particularly after turn 5. At that point, Annotator A assigns more wins to the witness, Annotator B fewer, and Annotator C none at all following the witness's admission to taking medication at turn 2—revealing an individual bias not shared by the others. Given the small number of annotators, we cannot draw definitive conclusions about NRA's suitability for this task, and leave for future work the question of whether NRA, or some refinement of it, has application to real-world text, following the approach of the small study in this section.

\section{Can LLMs Perceive Strategic Effects of Language?}
\begin{figure*}
    \centering
    \caption{Comparison of prompting techniques (GG: General guidelines; Few: Few-shot) on Qwen and QwQ, which perform well in the zero-shot setting. Few-shot prompting provides slight improvement, while Constitution prompting benefits PaT but harms BaT and NRBaT. All differences are minor ($< 0.1$), suggesting prompting has only a small or even negligible impact.}
    \includegraphics[width=1\linewidth]{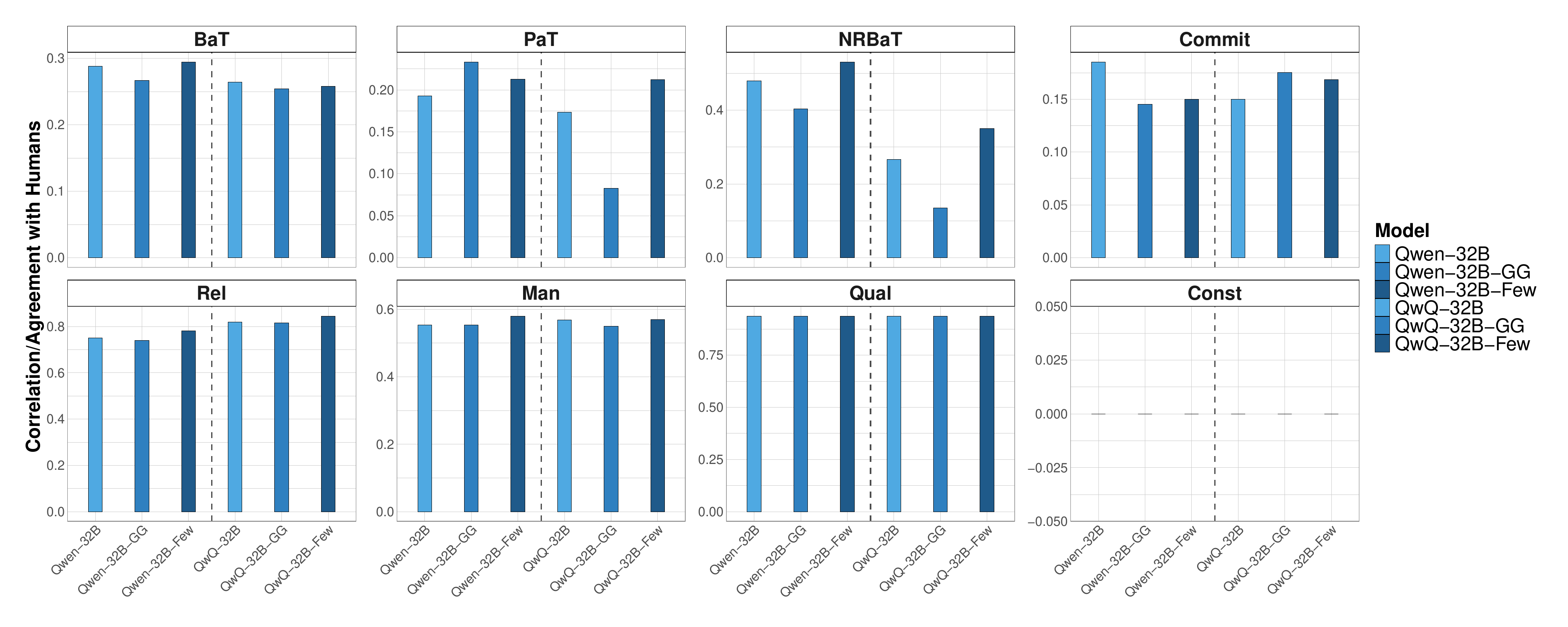}
    \vspace{-2em}
    \label{fig:qwen_comp}
\end{figure*}
% \begin{table}[t]
% \centering
% % \scriptsize
% \begin{tabular}{lll}
% \toprule
% \textbf{Base Model} & \textbf{Reasoning Variant}\\
% \midrule
% \texttt{GPT4o-mini} & \texttt{GPTo3-mini}\\
% \texttt{Gemini-2.5-Flash (OFF)} & \texttt{Gemini-2.5-Flash (ON)} \\
% \texttt{Qwen2.5-7B-Instruct} &  \texttt{DeepSeek-R1-Distill-Qwen-7B} \\
% \texttt{LLaMA3.1-Instruct-8B} &  \texttt{DeepSeek-R1-Distill-LLaMA-8B}\\
% \midrule
% \texttt{Qwen2.5-32B} & \texttt{QwQ-32B} \\
% \texttt{LLaMA3.3-70B-Instruct} & \texttt{DeepSeek-R1-Distill-LLaMA-70B}\\
% \bottomrule
% \end{tabular}

% \caption{Models Categorized by Size and Reasoning Capability; \texttt{Gemini-2.5-Flash (OFF)} refers to \texttt{gemini-2.5-flash-preview-05-20} with the thinking budget set to 0.}

% \label{Models}
% \end{table}

We evaluate a set of strong LLMs (Table~\ref{Models}), and further examine how model size and reasoning ability influence their perception of strategic dialogue. We use the same prompt provided to human annotators in a zero-shot setup. To constrain the variability of model responses, we set the temperature to 0.1. We provide the prompt in Appendix \ref{prompts}. 
We also experimented with different prompting techniques, including few-shot prompting and providing general guidelines for how to interpret the utterances, and found them largely leading to consistent results (see Figure \ref{fig:qwen_comp}). We emphasize that the analyses in this section are \emph{exploratory}. Our goal is to probe how current LLMs behave on strategic cross-examination dialogue, rather than to make strong claims about their underlying reasoning mechanisms.

\begin{wraptable}{r}{0.5\textwidth}
\begin{minipage}{\linewidth}
\centering
\scriptsize
\begin{tabular}{lll}
\toprule
\textbf{Instruct-tuned Model} & \textbf{Reasoning Variant}\\
\midrule
\texttt{GPT4o-mini} & \texttt{GPTo3-mini}\\
\texttt{Gemini-2.5-Flash (OFF)} & \texttt{Gemini-2.5-Flash (ON)} \\
\texttt{Qwen2.5-7B-Instruct} &  \texttt{DeepSeek-R1-Distill-Qwen-7B} \\
\texttt{LLaMA3.1-Instruct-8B} &  \texttt{DeepSeek-R1-Distill-LLaMA-8B}\\
\midrule
\texttt{Qwen2.5-32B-Instruct} & \texttt{QwQ-32B} \\
\texttt{LLaMA3.3-70B-Instruct} & \texttt{DeepSeek-R1-Distill-LLaMA-70B}\\
\bottomrule
\end{tabular}
\caption{Models Categorized by Size and Reasoning Capability; \texttt{Gemini-2.5-Flash (OFF)} refers to \texttt{gemini-2.5-flash-preview-05-20} with the thinking budget set to 0.}
\label{Models}
\end{minipage}
\vspace{-2em}
\end{wraptable} 

\subsection{Quantitative Analysis}
We report agreement/correlation with human annotations on \pat, \bat, \nrbat{} and also scores on their ``local'' benefit estimation components (i.e., \commit, \rel, \man, \qual, \const) aggregated across all three trials in Figure~\ref{fig:aggregated_results}, with detailed scores and significance levels for each trial in Appendix \ref{full_results}.

Overall, LLMs show strong agreement with humans in identifying violations of Gricean maxims, with  mean (denoted as $\mu$ hereafter) Randolph's $\kappa$ scores of 0.80, 0.52, and 0.93 for \rel, \man, and \qual, respectively. This aligns with prior findings \citep{hu-etal-2023-fine} suggesting that LLMs have a good pragmatic understanding of Gricean maxims. Another contributing factor may be the skewed distribution of violations (see Figure~\ref{fig:maxim}), which makes the task easier and inflates Randolph's $\kappa$. In contrast, LLMs perform poorly on commitment type identification (\commit($\mu$ = 0.14)) and our strategic metrics (\bat ($\mu$ = 0.23), \pat ($\mu$ = 0.13), and \nrbat ($\mu$ = 0.27)) all of which lag behind human inter-annotator agreement/correlation (see section~\ref{intrinsic eval}). Another interesting finding is that most models fail to reliably identify self-contradictory statements (\const), whereas small models tend to achieve higher true positive rates. However, due to the rarity of such cases in our dataset, we refrain from drawing strong conclusions.

\begin{figure*}[t]  
    \centering
    \caption{Strategic metrics and agreement with humans across three trials. BaT, PaT, NRBaT: Spearman's $\rho$; Commit: Fleiss' $\kappa$; Relevance, Manner, Quality: Randolph's $\kappa$; Consistency: true positive rate. ($N.B.$, Inconsistencies do not occur in every trial; when there are no inconsistent utterances, the true positive rate is naturally 0.)}
    \includegraphics[width=\textwidth]{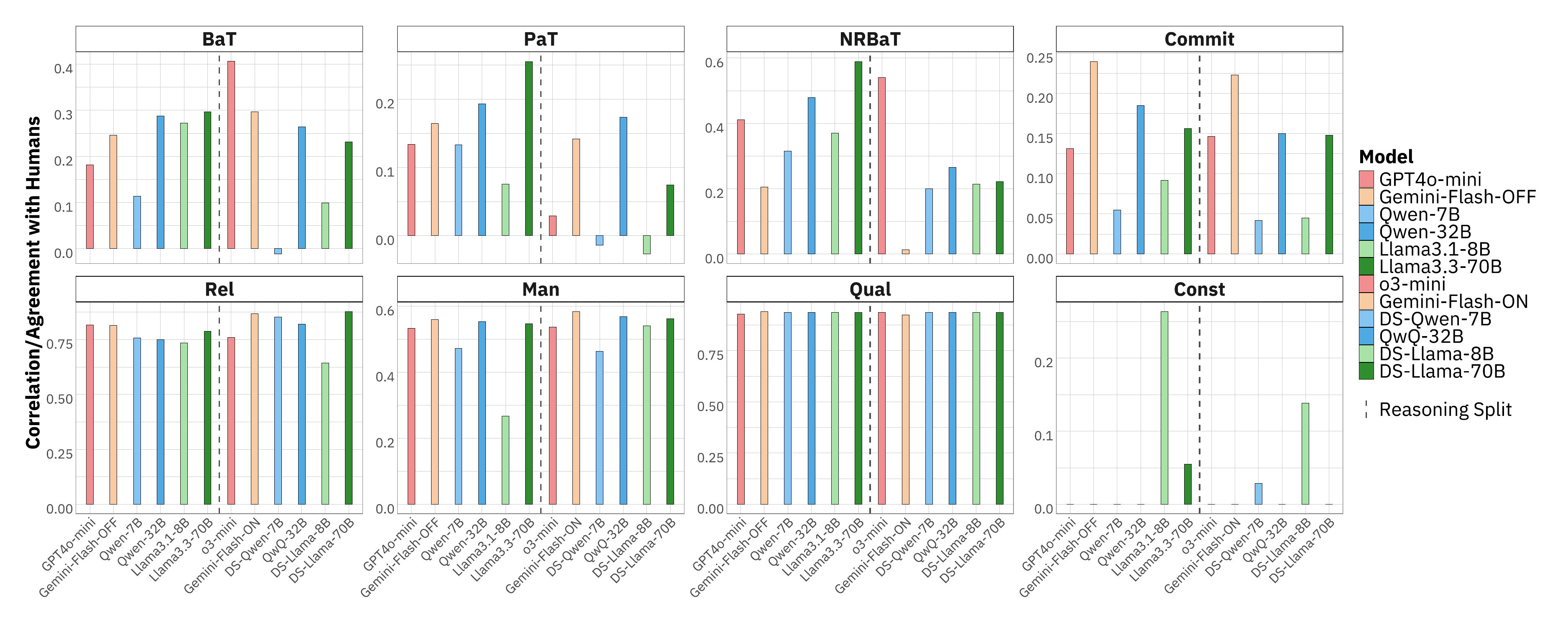}
    \vspace{-2em}
    \label{fig:aggregated_results}
    
\end{figure*}

\paragraph{Model size matters}
We find that larger models (indicated by darker bars in Figure~\ref{fig:aggregated_results}) consistently outperform their smaller counterparts (i.e., the lighter bars) on our BaT and PaT and in identifying commitment types. Below, we report average effect sizes ($\Delta\mu$) to quantify these differences, with bootstrapped 95\% confidence intervals (CI) for each metric; Bonferroni-corrected statistical significance from a paired t-test is denoted with a (*): \bat ($\Delta\mu=0.16$*, 95\% CI $[0.06, 0.25]$), \pat ($\Delta\mu = 0.12$, 95\% CI $[0.04, 0.22]$), and \commit ($\Delta\mu = 0.10$*, 95\% CI $[0.05, 0.15]$). The effects are inconsistent, though positive for \nrbat ($\Delta\mu = 0.08$, 95\% CI $[-0.07, 0.23]$), even when models exhibit improvements on BaT and PaT. This is because NRBaT is a cumulative measure that aggregates benefits across the entire discourse, making it less sensitive to the position of individual moves. Errors in local benefit estimation can cancel each other out over the course of the dialogue, which explains why models may sometimes exhibit low BaT and PaT scores but still achieve high NRBaT values, or vice versa.
Most models already perform well on violation identification, though we do observe that larger models tend to perform slightly better for example, on \man ($\Delta\mu = 0.12$*, 95\% CI $[0.06, 0.19]$).

\paragraph{Reasoning (CoT) does not help with our metrics} 
Models equipped with explicit reasoning mechanisms (i.e., the bars to the right of the dashed line in Figure \ref{fig:aggregated_results}) do not consistently improve performance and, in some cases, perform worse than their non-reasoning counterparts (i.e., bars to the left of the dashed line in the same color). This is particularly evident in \pat ($\Delta\mu = -0.10$*, 95\% CI $[-0.18, -0.04]$), and is also observed in \bat ($\Delta\mu = -0.03$, 95\% CI $[-0.11, 0.04]$) \nrbat ($\Delta\mu = -0.09$, 95\% CI $[-0.22, 0.03]$), and \commit ($\Delta\mu = -0.02$, 95\% CI $[-0.07, 0.02]$) across most models, with the exception of \texttt{GPT-o3-mini}. The confidence intervals for \bat\ and \commit\ include zero primarily due to this outlier. These results suggest that explicit reasoning tends to hinder models' ability to perceive strategic losses, if not benefits, as seen in the case of \texttt{GPT-o3-mini}. We further discuss possible reasons in section \ref{qual_analysis}. However, we do observe that models with reasoning ability outperform their counterparts on e.g., \man($\Delta\mu = 0.06$, 95\% CI $[0.02, 0.12]$) while the effects are weaker than model size. 
Lastly, we emphasize that our primary contribution lies in providing a method to probe how LLMs perceive the strategic effects of discourse moves. The results we present are restricted to the models examined under our prompting setup, and while they highlight certain issues (e.g., with CoT), we refrain from drawing strong conclusions in this fast-paced field.

\subsection{Qualitative Analysis} 
\label{qual_analysis}
We examine the sources of the observed effects of reasoning by analyzing the CoT that precedes generation, using models where such information is available: \texttt{Qwen}, \texttt{DS-LLaMA} and \texttt{Gemini-Flash}. While the reasoning traces provide a way to access explicit model judgements \citep{zaman2025chain}, we do not assume that these explanations faithfully reflect the model's internal decision process. Prior work shows that CoT traces may be unfaithful to model computation and can lead humans to infer incorrect narratives about ``how the model thinks'' \citep{10.5555/3666122.3669397, Levy2025HumansPW, Kambhampati2025StopAI}. Accordingly, we treat the generated reasoning as additional behavior to analyze, not as transparent access to the model's pragmatic representations.
\begin{wrapfigure}{r}{0.5\textwidth}
\begin{minipage}{\linewidth}
    \centering
    \includegraphics[width=\linewidth]{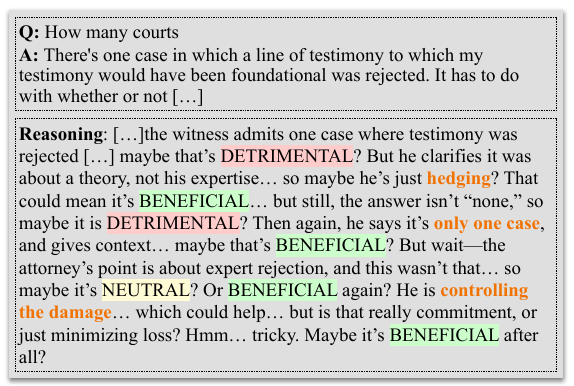}
    \caption{The model's reasoning confuses the detrimental commitment with loss-mitigation strategies.}
    \label{fig:qual_ex1}
\end{minipage}
\end{wrapfigure}
\paragraph{Reasoning traces involve overcomplication and ambiguity}

Models often overthink and conflate loss-minimizing or benefit-retrieval strategies with the actual impact of making a commitment. Consider a context as in Figure~\ref{fig:qual_ex1}; the prosecutor is asking whether the witness has ever been rejected as an expert in other courts. The response implies an affirmation, resulting in a detrimental commitment. While the model initially arrives at the correct judgment, it soon cycles through contradictory interpretations, ultimately settling on an incorrect assessment (i.e., \ben). The generated reasoning traces relies on surface features of the mitigation strategies without integrating their broader strategic implications. Moreover, the model sometimes misinterprets surface-level damage control strategies, such as minimizing, deflecting, or hedging (highlighted in orange), treating them in its reasoning traces as neutral or positive. This interpretation in the reasoning traces fails to recognize them as attempts to mitigate an otherwise detrimental commitment. We also append another example (Appendix~\ref{qual3}) where the model does the opposite (i.e., identifying benefit-retrieval strategy as detrimental). Misclassifying the type of commitment can strongly influence BaT and PaT scores by, e.g., inverting the base value. Even when the model correctly identifies violations, an incorrect commitment label can alter the interpretation of the violation itself, for example, leading the reasoning traces to construe a loss-control strategy (added to BaT) as if it were a benefit-retrieval strategy (added to PaT).

\paragraph{Reasoning traces are naive and sometimes contradictory}

\begin{wrapfigure}{r}{0.5\textwidth}
\begin{minipage}{\linewidth}
    \centering
    \includegraphics[width=\linewidth]{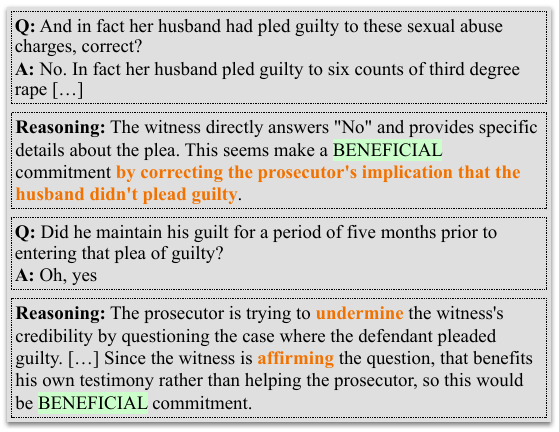}
    \caption{\small Example of model reasoning with misrepresentation of contextual information and cross-turn inconsistencies.}
    \label{fig:qual_ex2}
\end{minipage}
\end{wrapfigure}

%Models often \rev{produce reasoning traces that are} naive and sometimes contradictory, \rev{in the sense that 
The reasoning traces do not consistently reflect the discourse facts. In Figure~\ref{fig:qual_ex2}, the prosecutor is asking whether the defendant has pled guilty, admitting to which will be clearly detrimental to the defense side. The model's reasoning introduces a contrast not present in the context and shows misrepresentation of world information and contradicts itself across turns.
In the first exchange in the figure, the model claims the witness is correcting the idea that the husband didn't plead guilty, even though the question already presupposes that he did. 
%\rev{This mismatch reflects limitations of the explanation text rather than definitive evidence of how the model internally tracks the state of the world.} 
%By inventing a contrast that doesn't exist, the model \rev{appears, in its explanation, to overlook} the actual presuppositional structure, which \rev{in turn leads the explanation to} 

This mismatch is followed by a distorted
%distort its 
judgment of the witness's commitment. In several other cases, the reasoning traces focus primarily on whether the response affirms or rejects the question, rather than integrating the broader discourse or world knowledge. We avoid concluding that the model ``bases its judgment solely'' on polarity; instead, we note that this is one plausible interpretation of the observed reasoning traces under our current setup. The model's reasoning traces also exhibit self-contradiction in evaluating how a response aligns with the question's intent. For example, in the second exchange, it correctly recognizes that the prosecutor's question is designed to undermine the witness, yet it still judges the witness's affirmation as a beneficial commitment. More broadly, these patterns suggest that under our behavioral probes, the reasoning traces often align with surface-level informativeness rather than adversarial goals or commitment structure.
A more controlled experimental design would be required to determine the underlying causes of these behaviors. One promising direction is to draw on mechanistic interpretability methods \citep{geiger2025causalabstractionunderpinscomputational, huang2025internalcausalmechanismsrobustly, boguraev-etal-2025-causal}, which aim to identify internal mechanisms that contribute to specific model outputs. Another approach is to adopt controlled behavioral paradigms that vary the prompting context along minimal dimensions in order to probe what information LLMs rely on (e.g., \citealp{mccoy-etal-2019-right}). For example, one might manipulate only the polarity of a commitment, the presence of a pragmatic implicature, or the strategic role of the speaker. This makes it possible to test whether particular manipulations cause the model's judgments to shift in systematic ways, and hence to infer the causal role of each factor. Collectively, such techniques would allow for more principled conclusions about the extent and limitations of LLMs' internal strategic pragmatic competence.

\section{Discussions of Challenges and Limitations} 
\label{limitation}
Before closing, we discuss the implications of our proposal, the challenges it raises for extending 
the framework, and its practical limitations. In particular, we highlight methodological and 
conceptual limitations concerning annotation, scope of reasoning, and disciplinary grounding.

Although coherence and QUD share many similarities \citep{article},  they are not the same; hence using one notion rather than the other to tie between the discourse goals and moves may produce different predictions. 
We note that our treatment in terms of QUDs is in some respects stricter than coherence-based accounts. On many coherence-based theories, a response can be linguistically coherent so long as it maintains local discourse relations (for example, via entity links), even if it fails to address the current QUD.
%We note that our treatment via QUD is stricter than coherence-based accounts: responses that fail to address the QUD may still count as linguistically coherent under coherence-based theories (e.g., via entity links). 
The moves that we identify via QUDs as contributing to discourse goals are therefore a subset of those recognized under coherence-based theories. This has the implication that some coherence-based contributions are not recognized under our treatment. 

For example, consider (\ref{coherent}), a cross-examination from the Enron trial. It involves the witness Andrew Fastow (AF), the former Chief Financial Officer of Enron, who was called by the prosecution to testify about fraudulent practices within the company. The defense attorney (D) aims to challenge the reliability of his testimony.
The crucial part of the example is AF's response, where he does not commit to anything in response to the defense attorney's question. 

Under our QUD-based account, this response is therefore classified as making no meaningful contribution to the \emph{current} discourse goal, even though it remains linguistically coherent, for instance, under a relation such as ``follow-up question" \citep{li2022sdrt}. This illustrates how our notion of contribution is stricter than coherence-based accounts: some moves that are coherent are nevertheless treated as non-contributive with respect to the active QUD.

We acknowledge that our approach concerns more whether a turn advances the speaker's goals at the \textit{current} stage of the interaction and may miss aspects of its longer-term or global effects. Avoiding commitment now may serve longer-term purposes, such as delaying disclosure, or forcing redirecting the line of questioning. This distinction connects to our earlier discussion of short-term versus long-term strategic value and suggests that a comprehensive theory of strategic dialogue must ultimately account for both immediate goal advancement (as in the current paper) and deferred or indirect strategic effects, which we leave for future work.

\ea 
\label{coherent}
\begin{itemize}
    \item [D:] Now, you say, ``They stole in different ways," other members of senior management. What you're saying is that other members of senior management committed fraud to make their stock go up, then they would sell their stock and get away with the booty that way. That's what you're suggesting, right?
    \item [AF:] Are you asking me?
\end{itemize}
\z

A further implication concerns the simplification of credibility and the simplicity of the taxonomy of commitments. First, our decomposition of $\textnormal{P}_k(\textnormal{Good}_i)$ is not a theoretically faithful translation: violations of maxims do not by themselves constitute evidence of non-credibility as we have pointed out. The original term subsumes a broader range of factors, including, e.g., prior biases the jury may hold about the interlocutor. As a result, our function does not capture these more personal and socially grounded aspects of credibility as we have shown via experiments in section \ref{objective}, but only discourse-internal cues that can be operationalized through maxim violations. This makes our approach feasible but also narrower in scope compared to the original jury in ME Games.

In addition, our framework distinguishes only four commitment types and does not further subdivide the degrees of benefit or harm. As a result, a commitment that directly incriminates a witness could, in our framework, be weighted similarly to one with relatively minor negative implications. While a finer-grained taxonomy might capture such distinctions more accurately, our simplification serves as the first step to produce a function that is practically applicable.

We believe that these implications do not substantially undermine our proposal, though they point to directions where refinements could be made. Beyond these implications, a more ambitious challenge concerns how to operationalize constructs such as winning potential. While we do not tackle this challenge in the current paper, it would be essential for anyone aiming to develop a more accurate predictor of conversational outcomes. If one were to operationalize the term $\textnormal{win}_i(k)$, one would first need to identify the possible winning paths, which in turn requires a clear understanding of the speaker's conversational goals. In practice, however, speakers often pursue multiple goals of varying importance. While these goals can sometimes be informally described, as in (\ref{WMT}), there is no guarantee of a shared understanding of their relative importance. As a result, people are likely to have nuanced and divergent interpretations of a speaker's discourse goal hierarchy, and thus of whether those goals have become unattainable (i.e., whether winning potential has collapsed to zero).

Realizing $\textnormal{win}_i(k)$ in practice would therefore require (1) identifying the hierarchy of goals a speaker has, and (2) specifying how each goal can be achieved. Both are meaningful but ambitious undertakings that lie beyond the scope of this paper. We instead encourage readers interested in accurate outcome prediction to further investigate how this term might be operationalized.

Alongside these open challenges for future modeling, it is also important to recognize the practical limitations of our present study. Due to time and resource constraints, we were unable to annotate all the data we collected. Furthermore, the fact that neither annotators nor authors are legal scholars implies that we do not take into account strategies that require legal expertise to recognize, which we leave for future work. In addition, our annotations focus on local benefits: we did not require annotators to make long-horizon predictions. For example, a detrimental commitment was annotated as detrimental in the immediate context, even though it might ultimately serve a longer-term benefit (e.g., avoiding a later, more damaging commitment). Incorporating such long-term reasoning would require annotators to have a comprehensive understanding of the interlocutors' goal structures and would likely introduce greater subjectivity into the task. Therefore, we again leave this aspect to future research. A related limitation arises from annotating at the level of dialogue turns rather than finer-grained discourse segments. While turn-level annotation improves feasibility and consistency, it necessarily abstracts away from internal discourse structure: a single turn may simultaneously advance and undermine a speaker's strategic position, and more fine-grained segmentation could capture such mixed effects more precisely. Although our annotation scheme can still reflect this tension at the turn level, it does so only implicitly. Capturing such mixed strategic effects more faithfully would require a principled segmentation of sub-turn discourse units, a nontrivial methodological decision that lies beyond the scope of the present study.

A related limitation of our work is that, while we model trial discourse as a paradigmatic case of cooperative versus non-cooperative interaction, we have not yet fully connected our framework to the substantial literature in legal discourse analysis. Decades of research in conversation analysis and forensic linguistics have documented in detail how lawyers use questioning strategies to manage witnesses, control turn-taking, and shape what counts as relevant information \citep{atkinson1979order,drew1992contested,cotterill2003language,heffer2005language}. This work highlights the subtle interplay between institutional constraints, question design, and strategic dialogues. Our current study abstracts away from these details in order to test computational methods. In future work, we aim to integrate insights from courtroom discourse research, both to enrich the empirical grounding of our categories and to ensure that the strategic phenomena we identify align with what is already known about adversarial questioning practices.

A final limitation regards AI-safety and alignment issues. We acknowledge the potential safety concerns that come with computational modeling of sophisticated strategic interactions. For this reason, we do not pursue alignment questions directly in this work. Instead, we frame our contribution as a step toward providing principled tools for calibrating how models interpret and respond to strategic moves. Our hope is that such tools can complement future alignment research by supporting both empirical and theoretical investigations into whether AI agents can reason appropriately about non-cooperative discourse, and into whether agents that recognize non-cooperation nonetheless behave cooperatively. 

\section{Conclusions}
In this paper we introduced Strategic Dialogue Assessment (\name),  unifying two established pragmatic theories and providing a means to quantify their effects. Our successful application of \name\ to the Crooked Path Dataset demonstrates that \name\ has the potential to deepen our understanding of what makes discourse strategic. We see the current work as opening several promising directions for future research. First, although our analysis focuses on legal cross-examinations, the framework naturally extends to other high-stakes adversarial domains, such as political debates or negotiations. We feel there would be significant interest in any connection that could be drawn between such discourses and actual outcomes, for example via public opinion measures or voting outcomes for politicians, or quantifiable outcomes in negotiations. Second, a strategic agent reasons not only about how an utterance is perceived, but also about how other participants are likely to respond. This raises the question of how to model downstream reasoning (e.g., as in RSA), in which one interlocutor anticipates the other's decisions and strategically plans their own response. The current paper provides a first step in this direction by showing how to operationalize the utility function in adversarial settings. Building on this, predictive reasoning models such as RSA could incorporate strategic reasoning by optimizing \name\ jury over appropriately specified world states and lexicon. A very simple approach would be to optimize accumulated turn-level advantage, i.e., $\sum_{t=1}^{N} (\bat_t - \pat_t)$ ($N$ is the number of the moves), while leaving a more careful treatment of the constructs and additional potential hyperparameters (e.g., the weights for \bat\ and \pat) to future work.

\section*{Acknowledgments}
We are especially grateful to Kathryn Kazanas, Chenxing Riley Zhang, and the other annotators for their contributions to the data annotation for this project. We also thank Susan Yorke for generously sharing her legal expertise, which contributed to the legal accuracy and soundness of this work. We are grateful to the three anonymous \textit{D\&D} reviewers and the editor for their invaluable comments, which greatly strengthened the paper. Finally, we thank the members of the UT Computational Linguistics group for their insightful feedback and suggestions at various stages of this paper. This work was partially supported by NSF grants IIS-2145479, IIS-2107524, a grant from Open Philanthropy, and Good Systems,\footnote{\url{https://goodsystems.utexas.edu/}} a UT Austin Grand Challenge to develop responsible AI technologies. 
% \dibil{We should thank Susan Yorke? We should also thank 3 anonypous reviewers and the editor. And members of the UT CompLing group if they've heard presentations. Who else have we received comments from? }

\bibliography{dnd}

\newpage
\appendix
\section{Full GPT-4o response for Fig \ref{fig:illustration}}
\begin{figure}[H]
    \centering
    \includegraphics[width=0.9\linewidth]{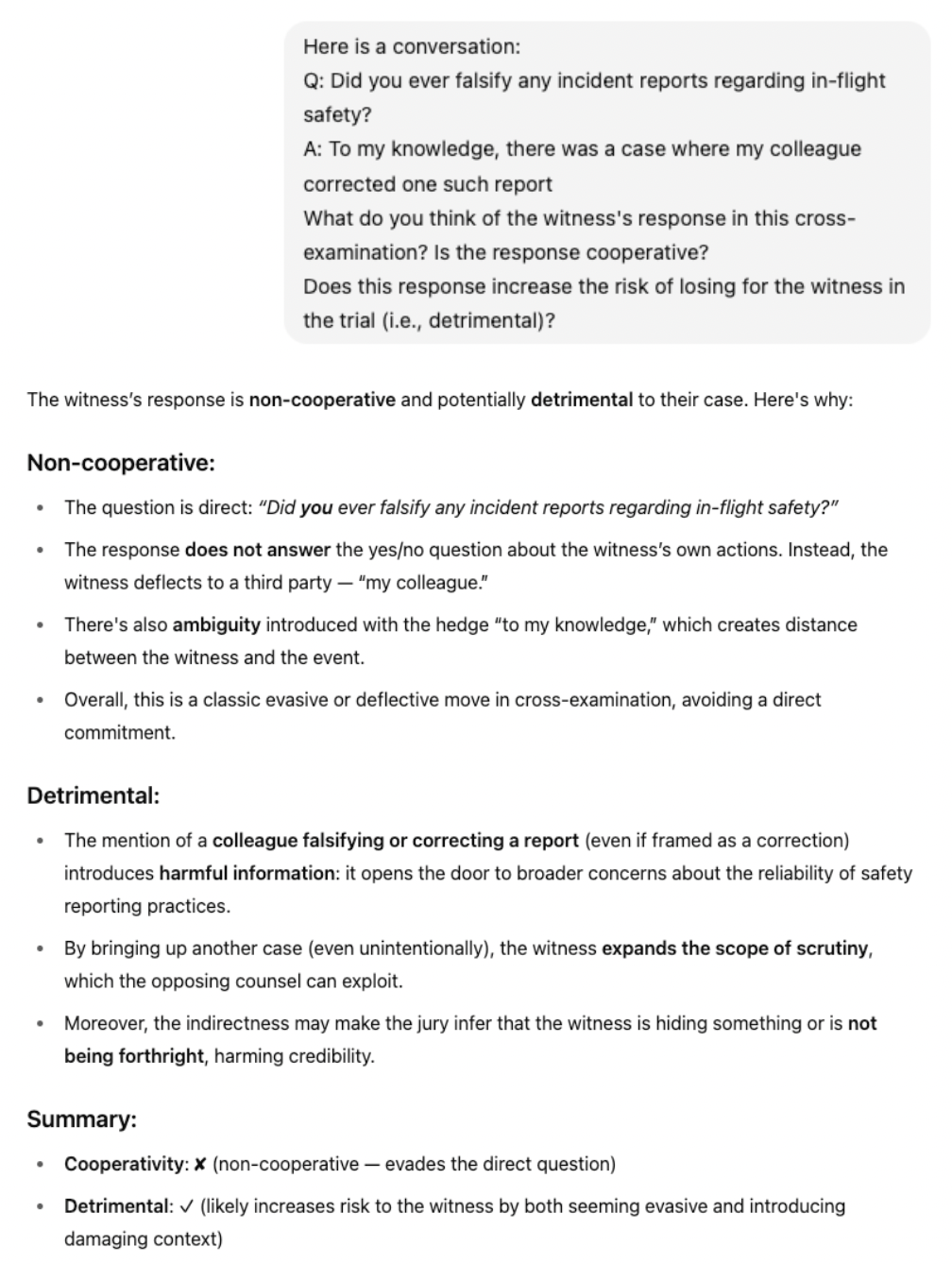}
    \caption{Full response from GPT-4o using the prompt in Fig \ref{fig:illustration}}
    \label{gptfull}
\end{figure}
\newpage
\section{Metrics Distribution in Human Annotations}
% \begin{table}[H]
% \centering
% \scriptsize
% \begin{tabular}{lcccc}
% \toprule
% \textbf{Trial} & \textbf{Defense} & \textbf{Prosecution} & \textbf{Total} & \textbf{Defense \%} \\ \midrule
% WMT     & 651 & 575 & 1226  & 53.1\% \\
% Enron   &  27 &  47 &  74  & 36.5\% \\
% Simpson & 1608 & 417 & 2025 & 79.4\% \\
% \bottomrule
% \end{tabular}
% \caption{Q/A pair distribution by questioner role across four trials. The Defense \% column shows the proportion of defense-attorney-led Q/A pairs.}
% \label{tab:qa-distribution}

% \end{table}

\begin{figure}[H]
    \centering
    \includegraphics[width=\linewidth]{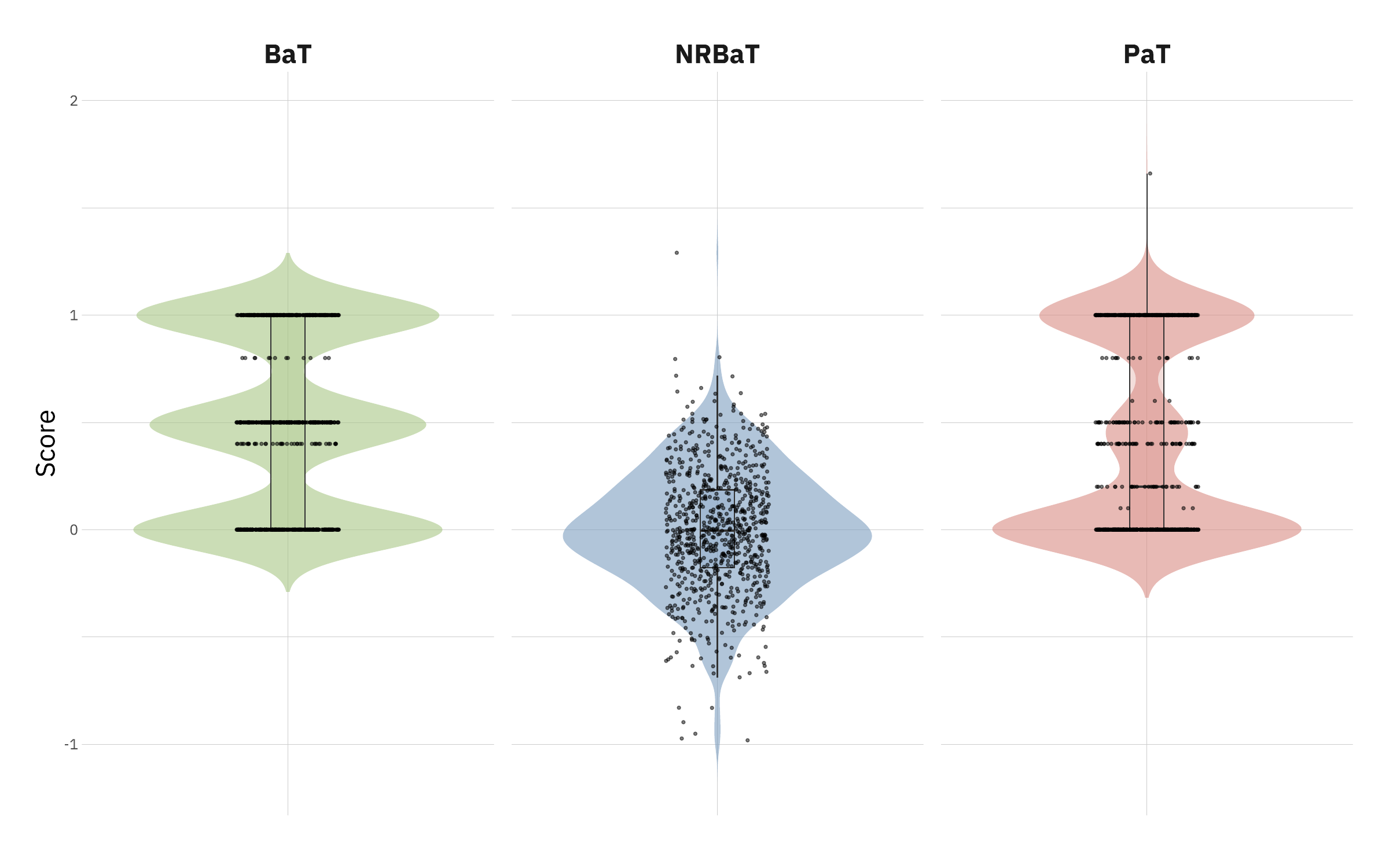}
    \caption{Distribution of our metrics in our human annotated dataset}
    \label{fig:enter-label}
\end{figure}

\newpage
\section{Metrics Calculation Example}
\label{cal_steps}
Take turn $i + 1$ in Figure~\ref{fig:calculation} as an example. The distributions of the BaT and PaT sums are (1, 1.4, 1.9, 1.9) and (0, 1, 1.2, 1.7), respectively, yielding $\mu_b = 0.98$, $\mu_p = 1.55$, $\sigma_b = 0.38$, and $\sigma_p = 0.14$ for the given example snippets. In our work, we assign weights of 0.4, 0.4, 0.2 and 0.2 to \rel, \man, \qual and \const, respectively. As noted in the main text, these weights are hypothetical, as our focus is on measuring correlation rather than modeling absolute values. The lower weight assigned to \qual/\const reflects its basis in perceived, rather than objective, truth; this prevents placing undue penalty on this dimension.
\begin{align*}
    \textnormal{BaT}_{i+1}    &= 
   |f_c(C_i)| \times (\textnormal{Rel}_{i+1} + \textnormal{Man}_{i+1} + \textnormal{Qual}_{i+1})   \\ &= 1 \times (0 + 0.4 + 0)\\ &= 0.4\\
    \textnormal{PaT}_{i+1}    &= |f_c(C_i)| + \textnormal{Const}_{i+1} \times \sum_{j=1}^{i+1}\textnormal{BaT}_j \\
                          &= 1 + 0 \\
                          &= 1\\
\textnormal{NRBaT}_{i+1}    &= Z\left(\sum_{j=1}^{i+1} \textnormal{BaT}_j\right) - Z\left(\sum_{j=1}^{i+1} \textnormal{PaT}_j\right)\nonumber\\
&= \frac{\sum_{j=1}^{i+1} \textnormal{BaT}_j - \mu_{b}}{\sigma_{b}} - \frac{\sum_{j=1}^{i+1} \textnormal{PaT}_j - \mu_{p}}{\sigma_{p}} \\
& = \frac{(0+1)-0.98}{0.38} - \frac{(1+0.4)-1.55}{0.14}\\
& = 0.053 - (-1.07)\\
& = 1.123
\end{align*}

\section{Zero-shot LLM prediction results}
\begin{table}[H]
\centering
\scriptsize
\begin{tabular}{llll}
\toprule
\textbf{Model} & \textbf{AUC}& \textbf{Model} & \textbf{AUC}\\
\midrule
GPT4o-mini        & 0.61 & o3-mini      & 0.60 \\
Gemini-Flash-OFF  & \textbf{0.68} & Gemini-Flash-ON   & 0.63 \\
Qwen-7B           & 0.53 & DS-Qwen-7B        & 0.47 \\
Qwen-32B          & 0.52 & QwQ-32B           & 0.58 \\
Llama3.1-8B       & 0.53 & DS-Llama-8B       & 0.51 \\
Llama3.3-70B      & 0.62 & DS-Llama-70B      & 0.59 \\
\bottomrule
\end{tabular}
\caption{AUC scores for 0-shot LLM predictions}
\label{tab:model_auc}
\end{table}
\newpage

\section{Qualitative Analysis Example}
\label{qual3}
% \noindent
% \begin{figure}[H]
%     \centering
    
%     \includegraphics[width=1\linewidth]{pics/qual_3.pdf}
    
%     \caption{An example of LLMs misidentifying benefit retrieval strategy as detrimental commitments.}
%     \label{fig:qual_ex3}
% \end{figure}
\codeboxinput[label=qual]{An example of LLMs misidentifying benefit retrieval strategy}{prompts/qual_ex.txt}

\section{Prompts}
\label{prompts}
\codeboxinput[label=prompt]{0-shot Prompt}{prompts/llm_prompt.txt}
\codeboxinput[label=cons]{General Guidelines}{prompts/constitution.txt}
\codeboxinput[label=few]{Few-shot examples}{prompts/few.txt}

\section{Annotation Protocols}
\label{protocol}
\codeboxinput[label=anno]{Annotation Instructions}{prompts/annotation_pro.txt}

\newpage
\section{Detailed Results}

\label{full_results}

\begin{table*}[t]
\centering
\scriptsize
\setlength{\tabcolsep}{10pt}
\renewcommand{\arraystretch}{0.5}

\begin{tabular}{lcccccccc}
\toprule
& \multicolumn{8}{c}{\textbf{Defense Witness vs. Prosecutor (WMT)}} \\
\midrule
\textbf{Model} & BaT & PaT & NRBaT & Commit & Rel & Man & Qual & $\textnormal{Const}^{\dagger}$ \\
\midrule
Human         & \phantom{-}0.65* & \phantom{-}0.66* & 0.83* & \phantom{-}0.59 & 0.72 & 0.52 & 0.86 & 0.25 \\
GPT4o-mini    & \phantom{-}0.27* & \phantom{-}0.11\phantom{*}  & 0.71* & \phantom{-}0.18 & 0.64 & 0.26 & 0.93 & 0 \\
Gemini-Flash-OFF    & \phantom{-}0.40* & \phantom{-}0.38*  & 0.62* & \phantom{-}0.38 & 0.70 & 0.34 & 0.93 & 0 \\
Qwen-7B       & \phantom{-}0.15* & \phantom{-}0.20* & 0.42* & \phantom{-}0.02 & 0.68 & 0.32 & 0.93 & 0 \\
Qwen-32B      & \phantom{-}0.32* & \phantom{-}0.21* & 0.69* & \phantom{-}0.19 & 0.71 & 0.29 & 0.93 & 0 \\
Qwen-32B-Few      & \phantom{-}0.33* & \phantom{-}0.23* & 0.75* & \phantom{-}0.20 & 0.72 & 0.29 & 0.93 & 0 \\
Qwen-32B-Cons      & \phantom{-}0.39* & \phantom{-}0.24* & 0.80* & \phantom{-}0.18 & 0.71 & 0.33 & 0.93 & 0 \\
LLaMA3.1-8B   & \phantom{-}0.02\phantom{*}  & \phantom{-}0.09\phantom{*}  & 0.12* & -0.03 & 0.64 & 0.19 & 0.93 & 0.58 \\
LLaMA3.3-70B   & \phantom{-}0.39*  & \phantom{-}0.25\phantom{*}  & 0.79* & 0.25 & 0.71 & 0.36 & 0.93 & 0.33 \\

\midrule
o3-mini       & \phantom{-}0.29* & \phantom{-}0.22* & 0.62* & \phantom{-}0.21 & 0.66 & 0.35 & 0.93 & 0 \\
Gemini-Flash-ON    & \phantom{-}0.46* & \phantom{-}0.44\phantom{*}  & 0.27* & \phantom{-}0.41 & 0.71 & 0.36 & 0.93 & 0 \\
DS-Qwen-7B    & -0.07\phantom{*} & -0.03\phantom{*} & 0.22* & \phantom{-}0.02 & 0.63 & 0.24 & 0.93 & 0.17 \\
QwQ-32B       & \phantom{-}0.21* & \phantom{-}0.17* & 0.59* & \phantom{-}0.14 & 0.72 & 0.36 & 0.93 & 0 \\
QwQ-32B-Few       & \phantom{-}0.33* & \phantom{-}0.31* & 0.71* & \phantom{-}0.27 & 0.74 & 0.40 & 0.93 & 0 \\
QwQ-32B-Cons      & \phantom{-}0.30* & \phantom{-}0.37* & 0.64* & \phantom{-}0.24 & 0.71 & 0.39 & 0.93 & 0 \\
DS-LLaMA-8B   & -0.08\phantom{*} & -0.05\phantom{*} & 0.11* & \phantom{-}0.05 & 0.66 & 0.43 & 0.93 & 0.83 \\
DS-LLaMA-70B  & \phantom{-}0.34* & \phantom{-}0.19* & 0.55* & \phantom{-}0.17 & 0.72 & 0.41 & 0.93 & 0 \\
\bottomrule
\end{tabular}
\caption{Strategic metrics and agreement with humans for \textbf{Defense Witness vs. Prosecutor (WMT)}. Stars (*) indicate significance at $p < .05$. BaT, PaT, NRBaT: Spearman's $\rho$; Commitment: Fleiss's $\kappa$; Relevance, Manner, Quality: Randolph's $\kappa$; Consistency: true positive rate ($N.B.$, Inconsistencies do not occur in every trial; when there are no inconsistent utterances, the true positive rate is naturally 0. We use $\dagger$ to indicate trials that do involve inconsistency.).}
\label{tab:defense-prosecutor}
\end{table*}

\begin{table*}[t]
\centering
\scriptsize
\setlength{\tabcolsep}{10pt}
\renewcommand{\arraystretch}{0.5}

\begin{tabular}{lcccccccc}
\toprule
& \multicolumn{8}{c}{\textbf{Prosecutor Witness vs. Defense (WMT)}} \\
\midrule
\textbf{Model} & BaT & PaT & NRBaT & Commit & Rel & Man & Qual & Const \\
\midrule
GPT4o-mini      & 0.09\phantom{*}  & 0.04\phantom{*}  & -0.24* & 0.16\phantom{*} & 0.72\phantom{*} & 0.64\phantom{*} & 0.98\phantom{*} & 0\phantom{*} \\
Gemini-Flash-OFF    & 0.06\phantom{*} & 0.10\phantom{*}  & \phantom{-}0.51* & 0.24\phantom{*} & 0.89\phantom{*} & 0.66\phantom{*} & 0.93\phantom{*} & 0\phantom{*} \\
Qwen-7B         & 0.07\phantom{*}  & 0.07\phantom{*}  & \phantom{-}0.62*  & 0.05\phantom{*} & 0.72\phantom{*} & 0.52\phantom{*} & 0.98\phantom{*} & 0\phantom{*} \\
Qwen-32B        & 0.25*            & 0.31*            & \phantom{-}0.30*  & 0.10\phantom{*} & 0.85\phantom{*} & 0.66\phantom{*} & 0.98\phantom{*} & 0\phantom{*} \\
Qwen-32B-Few        & 0.38*            & 0.43*            & \phantom{-}0.73*  & 0.19\phantom{*} & 0.84\phantom{*} & 0.66\phantom{*} & 0.98\phantom{*} & 0\phantom{*} \\
Qwen-32B-Cons        & 0.20*            & 0.22*            & \phantom{-}0.25*  & 0.13\phantom{*} & 0.82\phantom{*} & 0.66\phantom{*} & 0.98\phantom{*} & 0\phantom{*} \\
LLaMA3.1-8B    & 0.13\phantom{*}  & 0.16\phantom{*}  & \phantom{-}0.76*  & 0.07\phantom{*} & 0.76\phantom{*} & 0.33\phantom{*} & 0.98\phantom{*} & 0\phantom{*} \\
LLaMA3.3-70B    & 0.32*  & 0.28* & \phantom{-}0.83*  & 0.23\phantom{*} & 0.85\phantom{*} & 0.68\phantom{*} & 0.98\phantom{*} & 0\phantom{*} \\

\midrule
o3-mini         & 0.15\phantom{*}  & \phantom{-}0.02\phantom{*}  & 0.55*  & 0.18\phantom{*} & 0.84\phantom{*} & 0.61\phantom{*} & 0.98\phantom{*} & 0\phantom{*} \\
Gemini-Flash-ON    & 0.20* & \phantom{-}0.14\phantom{*}  & 0.24* & 0.33\phantom{*} & 0.90\phantom{*} & 0.66\phantom{*} & 0.98\phantom{*} & 0\phantom{*} \\
DS-Qwen-7B      & 0.06\phantom{*}  & \phantom{-}0.19*            & 0.70*  & 0.05\phantom{*} & 0.76\phantom{*} & 0.59\phantom{*} & 0.98\phantom{*} & 0\phantom{*} \\
QwQ-32B         & 0.36*            & \phantom{-}0.21*            & 0.48*  & 0.29\phantom{*} & 0.85\phantom{*} & 0.72\phantom{*} & 0.98\phantom{*} & 0\phantom{*} \\
QwQ-32B-Few        & 0.26*            & \phantom{-}0.20*            & 0.63*  & 0.19\phantom{*} & 0.82\phantom{*} & 0.64\phantom{*} & 0.98\phantom{*} & 0\phantom{*} \\
QwQ-32B-Cons       & 0.22*            & \phantom{-}0.14\phantom{*}            & 0.30*  & 0.21\phantom{*} & 0.82\phantom{*} & 0.66\phantom{*} & 0.98\phantom{*} & 0\phantom{*} \\
DS-LLaMA-8B     & 0.13\phantom{*}  & -0.03\phantom{*} & 0.73*  & 0.02\phantom{*} & 0.66\phantom{*} & 0.48\phantom{*} & 0.98\phantom{*} & 0\phantom{*} \\
DS-LLaMA-70B    & 0.24*            & \phantom{-}0.15\phantom{*}  & 0.59*  & 0.22\phantom{*} & 0.87\phantom{*} & 0.64\phantom{*} & 0.98\phantom{*} & 0\phantom{*} \\

\bottomrule
\end{tabular}
\label{tab:prosecution-witness}
\end{table*}

\begin{table*}[t]
\centering
\scriptsize
\setlength{\tabcolsep}{10pt}
\renewcommand{\arraystretch}{0.5}

\begin{tabular}{lcccccccc}
\toprule
& \multicolumn{8}{c}{\textbf{Defense Witness vs. Prosecutor (Enron)}} \\
\midrule
\textbf{Model} & BaT & PaT & NRBaT & Commit & Rel & Man & Qual & $\textnormal{Const}^{\dagger}$ \\
\midrule
GPT4o-mini       &  \phantom{-}0.10\phantom{*} &  \phantom{-}0.17\phantom{*} &  \phantom{-}0.14\phantom{*} &  \phantom{-}0.11\phantom{*} & \phantom{-}0.82\phantom{*} & \phantom{-}0.41\phantom{*} & \phantom{-}1.00\phantom{*} & 0 \\
Gemini-Flash-OFF    & \phantom{-}0.23* & -0.30*  & -0.71* & \phantom{-}0.13\phantom{*} & \phantom{-}0.66\phantom{*} & \phantom{-}0.40\phantom{*} & \phantom{-}0.83\phantom{*} & 0 \\
Qwen2.5-7B       &  \phantom{-}0.19\phantom{*} &  \phantom{-}0.21\phantom{*} &  \phantom{-}0.02\phantom{*} &  \phantom{-}0.18\phantom{*} & \phantom{-}0.66\phantom{*} & \phantom{-}0.45\phantom{*} & \phantom{-}0.83\phantom{*} & 0 \\
Qwen2.5-32B      &  \phantom{-}0.11\phantom{*} &  \phantom{-}0.24\phantom{*} &  \phantom{-}0.28\phantom{*} &  \phantom{-}0.31\phantom{*} & \phantom{-}0.70\phantom{*} & \phantom{-}0.36\phantom{*} & \phantom{-}0.83\phantom{*} & 0 \\
Qwen2.5-32B-Few      &  \phantom{-}0.17\phantom{*} &  \phantom{-}0.03\phantom{*} &  \phantom{-}0.13* &  \phantom{-}0.21\phantom{*} & \phantom{-}0.74\phantom{*} & \phantom{-}0.49\phantom{*} & \phantom{-}0.83\phantom{*} & 0 \\
Qwen2.5-32B-GG      &  \phantom{-}0.23\phantom{*} &  \phantom{-}0.16\phantom{*} &  \phantom{-}0.46* &  \phantom{-}0.24\phantom{*} & \phantom{-}0.57\phantom{*} & \phantom{-}0.40\phantom{*} & \phantom{-}0.83\phantom{*} & 0 \\
LLaMA3.1-8B      &  \phantom{-}0.23\phantom{*} &         -0.01\phantom{*}    &  \phantom{-}0.10\phantom{*} &  \phantom{-}0.31\phantom{*} & \phantom{-}0.70\phantom{*} & \phantom{-}0.49\phantom{*} & \phantom{-}0.83\phantom{*} & 1 \\
LLaMA3.1-70B      &  \phantom{-}0.16\phantom{*} &         -0.31*    &  -0.69* &  \phantom{-}0.08\phantom{*} & \phantom{-}0.53\phantom{*} & \phantom{-}0.49\phantom{*} & \phantom{-}0.83\phantom{*} & 0 \\
\midrule
o3-mini          &  \phantom{-}0.86*           &         -0.46*              &  \phantom{-}0.88*           &  \phantom{-}0.07\phantom{*} & \phantom{-}0.57\phantom{*} & \phantom{-}0.36\phantom{*} & \phantom{-}0.83\phantom{*} & 0 \\
Gemini-Flash-ON    & \phantom{-}0.16\phantom{*} & -0.33*  & -0.76* & \phantom{-}0.11\phantom{*} & 0.81 & 0.44 & 0.81 & 0 \\
DS-Qwen-7B       &  \phantom{-}0.14\phantom{*} &  \phantom{-}0.03\phantom{*} &         -0.21\phantom{*}    &         -0.02\phantom{*}    & \phantom{-}1.00\phantom{*} & \phantom{-}0.32\phantom{*} & \phantom{-}0.83\phantom{*} & 0 \\
QwQ-32B          &  \phantom{-}0.04\phantom{*} &  \phantom{-}0.11\phantom{*} &         -0.34*              &  \phantom{-}0.07\phantom{*} & \phantom{-}0.82\phantom{*} & \phantom{-}0.45\phantom{*} & \phantom{-}0.83\phantom{*} & 0 \\
QwQ-32B-Few          &  \phantom{-}0.07\phantom{*} &  -0.36* &         -0.68*              &  \phantom{-}0.09\phantom{*} & \phantom{-}0.95\phantom{*} & \phantom{-}0.49\phantom{*} & \phantom{-}0.83\phantom{*} & 0 \\
QwQ-32B-Cons          &  \phantom{-}0.25\phantom{*} &  -0.49* &         -0.79*              &  \phantom{-}0.11\phantom{*} & \phantom{-}0.74\phantom{*} & \phantom{-}0.36\phantom{*} & \phantom{-}0.83\phantom{*} & 0 \\
DS-LLaMA-8B     &  \phantom{-}0.25\phantom{*} &         -0.33*              &         -0.56*              &         -0.01\phantom{*}    & \phantom{-}0.49\phantom{*} & \phantom{-}0.28\phantom{*} & \phantom{-}0.83\phantom{*} & 0 \\
DS-LLaMA-70B     &  \phantom{-}0.25\phantom{*} &         -0.33*              &         -0.56*              &  \phantom{-}0.08\phantom{*} & \phantom{-}0.91\phantom{*} & \phantom{-}0.53\phantom{*} & \phantom{-}0.83\phantom{*} & 0 \\

\bottomrule
\end{tabular}

\label{tab:def-pros-enron}
\end{table*}

\begin{table*}[t]
\centering
\scriptsize
\setlength{\tabcolsep}{10pt}
\renewcommand{\arraystretch}{0.5}

\begin{tabular}{lcccccccc}
\toprule
& \multicolumn{8}{c}{\textbf{Prosecution Witness vs. Defense (Enron)}} \\
\midrule
\textbf{Model} & BaT & PaT & NRBaT & Commit & Rel & Man & Qual & Const \\
\midrule
GPT4o-mini       &  \phantom{-}0.30\phantom{*} &  \phantom{-}0.21\phantom{*} &  \phantom{-}0.90*           &  \phantom{-}0.08\phantom{*} & \phantom{-}0.91\phantom{*} & \phantom{-}0.53\phantom{*} & \phantom{-}0.78\phantom{*} & 0 \\
Gemini-Flash-OFF    & \phantom{-}0.32\phantom{*} & \phantom{-}0.35\phantom{*}  & \phantom{-}0.45* & \phantom{-}0.15\phantom{*} & 0.78 & 0.48 & 1.00 & 0 \\
Qwen2.5-7B       &  \phantom{-}0.18\phantom{*} &         -0.08\phantom{*}    &         -0.13\phantom{*}    &         -0.01\phantom{*}    & \phantom{-}0.78\phantom{*} & \phantom{-}0.33\phantom{*} & \phantom{-}1.00\phantom{*} & 0 \\

Qwen2.5-32B      &  \phantom{-}0.58*           &  \phantom{-}0.22\phantom{*} &  \phantom{-}0.77*           &  \phantom{-}0.34\phantom{*} & \phantom{-}0.41\phantom{*} & \phantom{-}0.63\phantom{*} & \phantom{-}1.00\phantom{*} & 0 \\
Qwen2.5-32B-Few      &  \phantom{-}0.38\phantom{*}        &  \phantom{-}0.31\phantom{*} &  \phantom{-}0.53*           &  \phantom{-}0.05\phantom{*} & \phantom{-}0.63\phantom{*} & \phantom{-}0.56\phantom{*} & \phantom{-}1.00\phantom{*} & 0 \\
Qwen2.5-32B-GG      &  \phantom{-}0.42\phantom{*}        &  \phantom{-}0.48* &  \phantom{-}0.37\phantom{*}            &  \phantom{-}0.11\phantom{*} & \phantom{-}0.56\phantom{*} & \phantom{-}0.56\phantom{*} & \phantom{-}1.00\phantom{*} & 0 \\

LLaMA3.1-8B      &  \phantom{-}0.10\phantom{*} &         -0.08\phantom{*}    &  \phantom{-}0.08\phantom{*} &  \phantom{-}0.08\phantom{*} & \phantom{-}0.48\phantom{*} & \phantom{-}0.11\phantom{*} & \phantom{-}1.00\phantom{*} & 0 \\

LLaMA3.3-70B      &  \phantom{-}0.52* &         0.77*    &  \phantom{-}0.97* &  \phantom{-}0.11\phantom{*} & \phantom{-}0.85\phantom{*} & \phantom{-}0.48\phantom{*} & \phantom{-}1.00\phantom{*} & 0 \\
\midrule
o3-mini          &  \phantom{-}0.38\phantom{*} &  \phantom{-}0.29\phantom{*} &  \phantom{-}0.19\phantom{*} &  \phantom{-}0.14\phantom{*} & \phantom{-}0.66\phantom{*} & \phantom{-}0.46\phantom{*} & \phantom{-}1.00\phantom{*} & 0 \\
Gemini-Flash-ON    & \phantom{-}0.43\phantom{*} & \phantom{-}0.34\phantom{*}  & \phantom{-}0.47* & \phantom{-}0.04\phantom{*} & \phantom{-}0.92\phantom{*} & \phantom{-}0.58\phantom{*} & \phantom{-}1.00\phantom{*} & 0 \\
DS-Qwen-7B       &         -0.07\phantom{*}    &         -0.34\phantom{*}    &         -0.19\phantom{*}    &  \phantom{-}0.19\phantom{*} & \phantom{-}1.00\phantom{*} & \phantom{-}0.41\phantom{*} & \phantom{-}1.00\phantom{*} & 0 \\
QwQ-32B          &  \phantom{-}0.44*           &  \phantom{-}0.18\phantom{*} &  \phantom{-}0.09\phantom{*} &         -0.02\phantom{*}    & \phantom{-}0.70\phantom{*} & \phantom{-}0.41\phantom{*} & \phantom{-}1.00\phantom{*} & 0 \\
QwQ-32B-Few          &  \phantom{-}0.39\phantom{*}          &  \phantom{-}0.54* &  \phantom{-}0.59* &         \phantom{-}0.06\phantom{*}    & \phantom{-}0.70\phantom{*} & \phantom{-}0.41\phantom{*} & \phantom{-}1.00\phantom{*} & 0 \\
QwQ-32B-Cons          &  \phantom{-}0.09\phantom{*}          &  \phantom{-}0.04\phantom{*} &  -0.05\phantom{*} &         -0.03\phantom{*}    & \phantom{-}0.70\phantom{*} & \phantom{-}0.41\phantom{*} & \phantom{-}1.00\phantom{*} & 0 \\
DS-LLaMA-8B      &  \phantom{-}0.05\phantom{*} &  \phantom{-}0.07\phantom{*} &  \phantom{-}0.21\phantom{*} &  \phantom{-}0.17\phantom{*} & \phantom{-}0.62\phantom{*} & \phantom{-}0.70\phantom{*} & \phantom{-}1.00\phantom{*} & 0 \\
DS-LLaMA-70B     &  \phantom{-}0.05\phantom{*} &  \phantom{-}0.07\phantom{*} &  \phantom{-}0.21\phantom{*} &  \phantom{-}0.11\phantom{*} & \phantom{-}0.93\phantom{*} & \phantom{-}0.41\phantom{*} & \phantom{-}1.00\phantom{*} & 0 \\

\bottomrule
\end{tabular}
\label{tab:pros-def-enron}
\end{table*}

\begin{table*}[t]
\centering
\scriptsize
\setlength{\tabcolsep}{10pt}
\renewcommand{\arraystretch}{0.5}

\begin{tabular}{lcccccccc}
\toprule
& \multicolumn{8}{c}{\textbf{Defense Witness vs. Prosecutor (Simpson)}} \\
\midrule
\textbf{Model} & BaT & PaT & NRBaT & Commit & Rel & Man & Qual & Const \\
\midrule
GPT4o-mini       &  \phantom{-}0.20*           &  \phantom{-}0.17*           &  \phantom{-}0.50*           &  \phantom{-}0.14\phantom{*} & \phantom{-}0.85\phantom{*} & \phantom{-}0.62\phantom{*} & \phantom{-}0.93\phantom{*} & 0 \\
Gemini-Flash-OFF    & \phantom{-}0.22* & \phantom{-}0.13\phantom{*}  & \phantom{-}0.12* & \phantom{-}0.24\phantom{*} & \phantom{-}0.89\phantom{*} & \phantom{-}0.69\phantom{*} & \phantom{-}0.93\phantom{*} & 0 \\
Qwen2.5-7B       &         -0.02\phantom{*}    &  \phantom{-}0.28*           &  \phantom{-}0.66*           &  \phantom{-}0.03\phantom{*} & \phantom{-}0.79\phantom{*} & \phantom{-}0.63\phantom{*} & \phantom{-}0.93\phantom{*} & 0 \\
Qwen2.5-32B      &  \phantom{-}0.17*           &  \phantom{-}0.13\phantom{*} &  \phantom{-}0.56*           &  \phantom{-}0.07\phantom{*} & \phantom{-}0.88\phantom{*} & \phantom{-}0.70\phantom{*} & \phantom{-}0.93\phantom{*} & 0 \\
Qwen2.5-32B-Few      &  \phantom{-}0.16\phantom{*}           &  \phantom{-}0.11\phantom{*} &  \phantom{-}0.52*           &  \phantom{-}0.12\phantom{*} & \phantom{-}0.83\phantom{*} & \phantom{-}0.76\phantom{*} & \phantom{-}0.93\phantom{*} & 0 \\
Qwen2.5-32B-GG      &  \phantom{-}0.14\phantom{*}           &  \phantom{-}0.11\phantom{*} &  \phantom{-}0.26*           &  \phantom{-}0.14\phantom{*} & \phantom{-}0.85\phantom{*} & \phantom{-}0.72\phantom{*} & \phantom{-}0.93\phantom{*} & 0 \\
LLaMA3.1-8B      &  \phantom{-}0.12\phantom{*} &  \phantom{-}0.17\phantom{*} &  \phantom{-}0.72*           &  \phantom{-}0.06\phantom{*} & \phantom{-}0.92\phantom{*} & \phantom{-}0.24\phantom{*} & \phantom{-}0.93\phantom{*} & 0 \\
LLaMA3.3-70B      &  \phantom{-}0.16\phantom{*} &  \phantom{-}0.21* &  \phantom{-}0.63*           &  \phantom{-}0.13\phantom{*} & \phantom{-}0.87\phantom{*} & \phantom{-}0.71\phantom{*} & \phantom{-}0.93\phantom{*} & 0 \\
\midrule
o3-mini          &  \phantom{-}0.30*           &  \phantom{-}0.06\phantom{*} &  \phantom{-}0.57*           &  \phantom{-}0.12\phantom{*} & \phantom{-}0.87\phantom{*} & \phantom{-}0.70\phantom{*} & \phantom{-}0.93\phantom{*} & 0 \\
Gemini-Flash-ON    & \phantom{-}0.20* & \phantom{-}0.04\phantom{*}  & -0.21* & \phantom{-}0.21\phantom{*} & \phantom{-}0.89\phantom{*} & \phantom{-}0.69\phantom{*} & \phantom{-}0.93\phantom{*} & 0 \\
DS-Qwen-7B       &         -0.15\phantom{*}    &         -0.01\phantom{*}    &  \phantom{-}0.44*           &         -0.05\phantom{*}    & \phantom{-}0.83\phantom{*} & \phantom{-}0.62\phantom{*} & \phantom{-}0.93\phantom{*} & 0 \\
QwQ-32B          &  \phantom{-}0.29*           &  \phantom{-}0.19*           &  \phantom{-}0.59*           &  \phantom{-}0.19\phantom{*} & \phantom{-}0.88\phantom{*} & \phantom{-}0.72\phantom{*} & \phantom{-}0.93\phantom{*} & 0 \\
QwQ-32B-Few          &  \phantom{-}0.19*           &  \phantom{-}0.27*           &  \phantom{-}0.62*           &  \phantom{-}0.20\phantom{*} & \phantom{-}0.90\phantom{*} & \phantom{-}0.71\phantom{*} & \phantom{-}0.93\phantom{*} & 0 \\
QwQ-32B-Cons          &  \phantom{-}0.25*           &  \phantom{-}0.23*           &  \phantom{-}0.48*           &  \phantom{-}0.26\phantom{*} & \phantom{-}0.86\phantom{*} & \phantom{-}0.73\phantom{*} & \phantom{-}0.93\phantom{*} & 0 \\
DS-LLaMA-8B      &  \phantom{-}0.10\phantom{*} &  \phantom{-}0.14\phantom{*} &  \phantom{-}0.59*           &  \phantom{-}0.05\phantom{*} & \phantom{-}0.64\phantom{*} & \phantom{-}0.66\phantom{*} & \phantom{-}0.93\phantom{*} & 0 \\
DS-LLaMA-70B     &  \phantom{-}0.29*           &  \phantom{-}0.24*           &  \phantom{-}0.54*           &  \phantom{-}0.17\phantom{*} & \phantom{-}0.88\phantom{*} & \phantom{-}0.70\phantom{*} & \phantom{-}0.93\phantom{*} & 0 \\

\bottomrule
\end{tabular}

\label{tab:def-pros-simpson}
\end{table*}

\begin{table*}[t]
\centering
\scriptsize
\setlength{\tabcolsep}{10pt}
\renewcommand{\arraystretch}{0.5}

\begin{tabular}{lcccccccc}
\toprule
& \multicolumn{8}{c}{\textbf{Prosecution Witness vs. Defense (Simpson)}} \\
\midrule
\textbf{Model} & BaT & PaT & NRBaT & Commit & Rel & Man & Qual & Const \\
\midrule
GPT4o-mini       &  \phantom{-}0.12\phantom{*} &  \phantom{-}0.10\phantom{*} &         -0.18*              &  \phantom{-}0.12\phantom{*} & \phantom{-}0.96\phantom{*} & \phantom{-}0.74\phantom{*} & \phantom{-}0.95\phantom{*} & 0 \\
Gemini-Flash-OFF    & \phantom{-}0.23* & \phantom{-}0.30*  & \phantom{-}0.24* & \phantom{-}0.30\phantom{*} & \phantom{-}0.96\phantom{*} & \phantom{-}0.79\phantom{*} & \phantom{-}0.95\phantom{*} & 0 \\

Qwen2.5-7B       &  \phantom{-}0.11\phantom{*} &  \phantom{-}0.11\phantom{*} &  \phantom{-}0.10\phantom{*} &  \phantom{-}0.06\phantom{*} & \phantom{-}0.91\phantom{*} & \phantom{-}0.58\phantom{*} & \phantom{-}0.95\phantom{*} & 0 \\
Qwen2.5-32B      &  \phantom{-}0.24*           &  \phantom{-}0.04\phantom{*} &  \phantom{-}0.03\phantom{*} &  \phantom{-}0.10\phantom{*} & \phantom{-}0.95\phantom{*} & \phantom{-}0.68\phantom{*} & \phantom{-}0.95\phantom{*} & 0 \\
Qwen2.5-32B-Few      &  \phantom{-}0.33*           &  \phantom{-}0.14\phantom{*} &  \phantom{-}0.33* &  \phantom{-}0.13\phantom{*} & \phantom{-}0.94\phantom{*} & \phantom{-}0.72\phantom{*} & \phantom{-}0.95\phantom{*} & 0 \\
Qwen2.5-32B-GG      &  \phantom{-}0.20*           &  \phantom{-}0.16* &  \phantom{-}0.06\phantom{*} &  \phantom{-}0.07\phantom{*} & \phantom{-}0.93\phantom{*} & \phantom{-}0.65\phantom{*} & \phantom{-}0.95\phantom{*} & 0 \\
LLaMA3.1-8B      &  \phantom{-}0.79*           &  \phantom{-}0.12\phantom{*} &  \phantom{-}0.13\phantom{*} &  \phantom{-}0.06\phantom{*} & \phantom{-}0.91\phantom{*} & \phantom{-}0.24\phantom{*} & \phantom{-}0.95\phantom{*} & 0 \\
LLaMA3.3-70b      &  \phantom{-}0.19*           &  \phantom{-}0.11\phantom{*} &  \phantom{-}-0.18* &  \phantom{-}0.14\phantom{*} & \phantom{-}0.92\phantom{*} & \phantom{-}0.56\phantom{*} & \phantom{-}0.95\phantom{*} & 0 \\
\midrule

o3-mini          &  \phantom{-}0.13\phantom{*} &  \phantom{-}0.07\phantom{*} &  \phantom{-}0.07\phantom{*} &  \phantom{-}0.16\phantom{*} & \phantom{-}0.95\phantom{*} & \phantom{-}0.74\phantom{*} & \phantom{-}0.95\phantom{*} & 0 \\
Gemini-Flash-ON    & \phantom{-}0.30* & \phantom{-}0.19*  & \phantom{-}0.25* & \phantom{-}0.24\phantom{*} & \phantom{-}0.98\phantom{*} & \phantom{-}0.77\phantom{*} & \phantom{-}0.95\phantom{*} & 0 \\
DS-Qwen-7B       &  \phantom{-}0.02\phantom{*} &  \phantom{-}0.09\phantom{*} &  \phantom{-}0.06\phantom{*} &  \phantom{-}0.06\phantom{*} & \phantom{-}0.89\phantom{*} & \phantom{-}0.60\phantom{*} & \phantom{-}0.95\phantom{*} & 0 \\
QwQ-32B          &  \phantom{-}0.22*           &  \phantom{-}0.18*           &  \phantom{-}0.02\phantom{*} &  \phantom{-}0.23\phantom{*} & \phantom{-}0.95\phantom{*} & \phantom{-}0.75\phantom{*} & \phantom{-}0.95\phantom{*} & 0 \\
QwQ-32B-Few          &  \phantom{-}0.29*           &  \phantom{-}0.26*           & -0.01\phantom{*} &  \phantom{-}0.20\phantom{*} & \phantom{-}0.96\phantom{*} & \phantom{-}0.77\phantom{*} & \phantom{-}0.95\phantom{*} & 0 \\
QwQ-32B-Few          &  \phantom{-}0.40*           &  \phantom{-}0.26*           & \phantom{-}0.33\phantom{*} &  \phantom{-}0.26\phantom{*} & \phantom{-}0.96\phantom{*} & \phantom{-}0.75\phantom{*} & \phantom{-}0.95\phantom{*} & 0 \\
DS-LLaMA-8B      &  \phantom{-}0.14\phantom{*} &  \phantom{-}0.05\phantom{*} &  \phantom{-}0.01\phantom{*} &         -0.01\phantom{*}    & \phantom{-}0.79\phantom{*} & \phantom{-}0.69\phantom{*} & \phantom{-}0.95\phantom{*} & 0 \\
DS-LLaMA-70B     &  \phantom{-}0.21*           &  \phantom{-}0.13*           &         -0.13\phantom{*}    &  \phantom{-}0.14\phantom{*} & \phantom{-}0.95\phantom{*} & \phantom{-}0.68\phantom{*} & \phantom{-}0.95\phantom{*} & 0 \\

\bottomrule
\end{tabular}

\label{tab:pros-def-simpson}
\end{table*}

\begin{table*}[t]
\centering
\scriptsize
\setlength{\tabcolsep}{10pt}
\renewcommand{\arraystretch}{0.5}

\begin{tabular}{lrrrrrrrr}
\toprule
\textbf{Metric} & \textbf{Wins} & \textbf{Loses} & \textbf{Ties} & \textbf{Mean} & \textbf{Median} & \textbf{SD} & \textbf{CI Low} & \textbf{CI High} \\
\midrule
BaT     & 18 & 4  & 2 & 0.16 & 0.19 & 0.24 & 0.06 & 0.25  \\
PaT     & 18 & 4  & 2 & 0.12 & 0.09 & 0.22 & 0.04 & 0.22  \\
Commit  & 21 & 3  & 0 & 0.10 & 0.12 & 0.13 & 0.05 & 0.15  \\
NRBaT   & 10 & 12 & 2 & 0.08 & -0.02 & 0.39 & -0.07 & 0.23 \\
Man     & 17 & 5  & 2 & 0.12 & 0.13 & 0.17 & 0.06 & 0.19  \\
Rel     &19 & 5 & 0 & 0.06 & 0.07 & 0.19 & -0.01 & 0.13 \\
Qual     & 0 & 1  & 23 & -8.33\text{e}{-4}
 & 0 & 4.08\text{e}{-3} & -2.5\text{e}{-3}
 & 0  \\

\bottomrule
\end{tabular}
\caption{Comparison by metric across models of different sizes. Wins indicate instances where the larger model outperforms the smaller one, while Loses indicate the opposite.}
\label{tab:comparison_metrics}
\end{table*}

\begin{table*}[t]
\centering
\scriptsize
\setlength{\tabcolsep}{10pt}
\renewcommand{\arraystretch}{0.5}

\begin{tabular}{lrrrrrrrr}
\toprule
\textbf{Metric} & \textbf{Wins} & \textbf{Loses} & \textbf{Ties} & \textbf{Mean} & \textbf{Median} & \textbf{SD} & \textbf{CI Low} & \textbf{CI High} \\
\midrule
BaT     & 12 & 17 & 1  & -0.03 & -0.02 & 0.22 & -0.11 & 0.05 \\
PaT     &  8 & 22 & 0  & -0.10 & -0.05 & 0.20 & -0.18 & -0.04 \\
NRBaT   & 10 & 20 & 0  & -0.09 & -0.08 & 0.36 & -0.22 & 0.03 \\
Commit  & 11 & 13 & 6  & -0.02 &  0 & 0.13 & -0.07 & 0.03 \\
Man     & 18 & 11 & 1  &  0.06 &  0.05 & 0.18 &  0 & 0.12 \\
Rel     & 18 &  9 & 3  &  0.02 &  0.02 & 0.16 & -0.04 & 0.08 \\
Qual    &  1 &  2 & 27 &  0 &  0 & 0.05 & -0.02 & 0.02 \\
\bottomrule
\end{tabular}

\caption{Comparison by metric across models with and without reasoning ability. Wins indicate instances where the reasoning model outperforms its non-reasoning counterpart, while Loses indicate the opposite.}
\label{tab:comparison_metrics_reasoning}
\end{table*}

\end{document}